\documentclass[
12pt, 
oneside, 
english, 
onehalfspacing, 
nolistspacing, 
parskip, 
headsepline, 
]{MastersDoctoralThesis} 

\usepackage[utf8]{inputenc} 
\usepackage[T1]{fontenc} 

\usepackage{mathpazo} 
\usepackage{comment}
\usepackage{amsmath}
\usepackage{graphicx}
\usepackage{subfig}
\DeclareCaptionOption{parskip}[]{} 
\usepackage{caption}

\usepackage[backend=bibtex,style=authoryear,natbib=true]{biblatex} 

\DeclareBibliographyCategory{cited}
\AtEveryCitekey{\addtocategory{cited}{\thefield{entrykey}}}

\nocite{*}
\usepackage[autostyle=true]{csquotes} 

\thesistitle{Spatial Attention as an Interface for Image Captioning Models}
\supervisor{Dr. Tatjana \textsc{Scheffler}} 
\examiner{Dr. David \textsc{Schlangen}} 
\degree{Master of Science in Cognitive Systems} 
\author{Philipp \textsc{Sadler}} 
\addresses{} 

\subject{} 
\university{University of Potsdam} 
\department{Department of Linguistics} 
\faculty{Faculty of Human Sciences} 

\AtBeginDocument{
\hypersetup{pdftitle=\ttitle} 
\hypersetup{pdfauthor=\authorname} 
\hypersetup{pdfkeywords=\keywordnames} 
}

\begin{document}

\frontmatter 

\pagestyle{plain} 


\begin{titlepage}
\begin{center}

\vspace*{.06\textheight}
{\scshape\LARGE \univname\par}\vspace{1.5cm} 
\textsc{\Large Master Thesis}\\[0.5cm] 

\HRule \\[0.4cm] 
{\huge \bfseries \ttitle\par}\vspace{0.4cm} 
\HRule \\[1.5cm] 
 
\begin{minipage}[t]{0.4\textwidth}
\begin{flushleft} \large
\emph{Author:}\\
{\authorname} 
\end{flushleft}
\end{minipage}
\begin{minipage}[t]{0.4\textwidth}
\begin{flushright} \large
\emph{1st Supervisor:} \\
{\supname} 
\end{flushright}
\begin{flushright} \large
	\emph{2st Supervisor:} \\
	{\examname} 
\end{flushright}
\end{minipage}\\[2cm]
 
\vfill

\large \textit{A thesis submitted in fulfillment of the requirements\\ for the degree \degreename}\\[0.3cm] 
 
\vfill

{\large \today}\\[4cm] 

\vfill
\end{center}
\end{titlepage}


\begin{declaration}
\noindent I, \authorname, declare that this thesis titled, \enquote{\ttitle} and the work presented in it are my own. I confirm that:

\begin{itemize} 
\item This work was done wholly or mainly while in candidature for a research degree at this University.
\item Where any part of this thesis has previously been submitted for a degree or any other qualification at this University or any other institution, this has been clearly stated.
\item Where I have consulted the published work of others, this is always clearly attributed.
\item Where I have quoted from the work of others, the source is always given. With the exception of such quotations, this thesis is entirely my own work.
\item I have acknowledged all main sources of help.
\item Where the thesis is based on work done by myself jointly with others, I have made clear exactly what was done by others and what I have contributed myself.\\
\end{itemize}
 
\noindent Signed:\\
\rule[0.5em]{25em}{0.5pt} 
 
\noindent Date:\\
\rule[0.5em]{25em}{0.5pt} 
\end{declaration}

\cleardoublepage






\begin{abstract}

The internal workings of modern deep learning models stay often unclear to an external observer, although spatial attention mechanisms are involved. The idea of this work is to translate these spatial attentions into natural language to provide a simpler access to the model's function. Thus, I took a neural image captioning model and measured the reactions to external modification in its spatial attention for three different interface methods: a fixation over the whole generation process, a fixation for the first time-steps and an addition to the generator's attention. The experimental results for bounding box based spatial attention vectors have shown that the captioning model reacts to method dependent changes in up to 52.65\% and includes in 9.00\% of the cases object categories, which were otherwise unmentioned. Afterwards, I established such a link to a hierarchical co-attention network for visual question answering by extraction of its word, phrase and question level spatial attentions. Here, generated captions for the word level included details of the question-answer pairs in up to 55.20\% of the cases. This work indicates that spatial attention seen as an external interface for image caption generators is an useful method to access visual functions in natural language. 

\end{abstract}

\begin{center}
	{\normalsize \MakeUppercase{Universität Potsdam} \par}
	\bigskip
	{\huge\textit{Zusammenfassung} \par}
	\bigskip
	{\normalsize Humanwissenschaftliche Fakultät \par}
	{\normalsize Department für Linguistik \par}
	\bigskip
	{\normalsize \degreename\par}
	\bigskip
	{\normalsize\bfseries \ttitle \par}
	\medskip
	{\normalsize \byname{} \authorname \par}
	\bigskip
\end{center}
{
	Für einen aussenstehenden Beobachter bleiben die internen Vorgänge eines neuronalen Netzwerkes häufig unklar, auch wenn Methoden der räumlichen Aufmerksamkeit involviert sind. Die Idee dieser Arbeit ist die Übersetzung dieser räumlichen Aufmerksamkeit in natürliche Sprache, um einen einfach-eren Zugang zu den Modellen zu eröffnen. Ich habe einen standardmäßigen Bildbeschreibungsgenerator genommen und seine Reaktionen bezüglich Manipulationen in seiner räumlichen Aufmerksamkeit für drei verschiedene Verfahren gemessen: eine Fixierung während des gesamten Generierungs-prozesses, eine Fixierung für die ersten Zeitschritte und eine Addition zu der Aufmerksamkeit des Generators. Die experimentellen Resultate für räumliche Aufmerksameit in Objektrahmen haben gezeigt, dass der Generator methodisch-spezifische Reaktionen in 52.65\% zeigt und in 9.00\% der Fälle Objekt Kategorien erwähnt, die vorher unerwähnt blieben. Anschließend stellte ich eine ebensolche Verbindung mit einem hierarchischen Ko-Aufmerk-samkeitsnetzwerkes her, indem ich die räumliche Aufmerksamkeit für Wort-, Phrase- und Fragestufe extrahierte. Hier beinhalteten 55.20\% der auf Wortstufen basierenden generierten Beschreibungen, Wörter der Frage-Antwort Paare. Diese Arbeit zeigt an, dass räumliche Aufmerksamkeit, gesehen als eine externe Schnittstelle für Bildbeschreibungsgeneratoren, eine nützliche Methode darstellt, um visuelle Funktionen sprachlich zugänglich zu machen.
	\vfill\null
}




\tableofcontents 

\listoffigures 

\listoftables 


\begin{abbreviations}{ll} 

\textbf{BLEU} & \textbf{B}i-\textbf{L}ingual \textbf{E}valuation \textbf{U}nderstudy\\
\textbf{CNN} & \textbf{C}onvolutional \textbf{N}eural \textbf{N}etwork\\
\textbf{LSTM} & \textbf{L}ong \textbf{S}hort-\textbf{T}erm \textbf{M}emory Network\\
\textbf{MSCOCO} & \textbf{M}icro\textbf{s}oft \textbf{Co}mmon Objects in \textbf{Co}ntext\\
\textbf{NLTK} & \textbf{N}atural \textbf{L}anguage \textbf{T}ool\textbf{K}it\\
\textbf{ReLU} & \textbf{Re}ctified \textbf{L}inear \textbf{U}nit\\
\textbf{RNN} & \textbf{R}ecurrent \textbf{N}eural \textbf{N}etwork\\
\textbf{R-CNN} & \textbf{R}egion Proposal \textbf{CNN}\\
\textbf{VQA} & \textbf{V}isual \textbf{Q}uestion \textbf{A}nswering\\

\end{abbreviations}















\mainmatter 

\pagestyle{thesis} 


\chapter{Introduction} 

\label{Chapter1} 

The modern interdisciplinary fields of natural language processing yield new challenges by involving both textual and visual data. Such a challenging task is visual question answering (VQA) which has been initially introduced as a Turing test proxy \parencite{malinowski_multi-world_2014}. A deep learning model has to provide an answer given both an image and a question about it \parencite{agrawal_vqa:_2017,zhu_visual7w:_2015}. 
These deep learning network architectures are roughly dividable into three sub-systems. There is the vision component that extracts useful information from the images, which is usually an already trained vision model. Then there is a probably also pre-trained language component that is learning a useful representation for the textual questions. Finally, there is the answer generator, which is combining both visual and textual language representation to generate an answer for the question.


In state-of-the-art architectures for this task, the focus is either set to learn a multi-modal high dimensional space for questions and images \parencite{fukui_multimodal_2016} or to utilize the textual input to align with specific visual stimuli given the images \parencite{leibe_ask_2016}. While the first approaches achieve higher scores, the latter allow a more intuitive interpretation of the models internal workings.
More intuitive in a sense that the model's chosen visual stimuli, what I call \textit{focus pattern} here, are presentable along with the given image. Then a human inspector is able to visually analyze the model's focus patterns given the answers. Although these are not necessarily enough evidence for the answer choice \parencite{serrano_is_2019}, still these focus patterns allow a visual and in such a sense, natural access to the model's decision making.

\clearpage



Nevertheless, the normal visual question answering task is only about the resulting answer. Another interesting access to the internal workings of the model would be to additionally expose the intermediate steps until answer conclusion in natural language. The idea is to use an image caption generator to directly convert the visual focus patterns into natural language statements.


The goal for the image captioning task is to generate a short description for a given image. This is usually done by letting the model learn where to look in the image, for example to detect the main objects and the general setting. Yet again, such a model is producing focus patterns to detect these visual entities which influence the model's word choice for the description.


Now, when the answer generator for a visual question answering model is relying on focus patterns to produce an answer, then these focus patterns are expressable in natural language using an image generator which also relies on such a visual processing. As a result, the internal workings respectively the intermediate steps of the answer generator are representable in human language.
In such a sense, the model would be enabled to express itself in a more human understandable way. Furthermore, these connection to an image captioning model is helpful in a variety of ways, for example:

\begin{itemize}
	\item Previous work showed that the textual question alone is already a strong predictor for the answer \parencite{ren_exploring_2015,zhu_visual7w:_2015}. With the idea above, an answer generator could solely rely on the image and the intermediately generated image descriptions instead of directly concluding an answer from the question. 
	\item The visual question answering is often formulated as a classification task, where only the most common answers are predicted. As such, the model is constrained to express itself from the answer set, even tough a richer vocabulary is necessary to interpret the questions. Given the idea above, the model would be enabled to provide additional human readable hints along with the chosen answer.
\end{itemize}

\clearpage


The expectation is that the images translated to brief descriptions provide a representative textual context for the questions like in natural language question answering tasks, act as a human readable explanation and extend the model's vocabulary for the answer selection. In particular, such a sub-task allows separate training of both networks and thus the usage of a larger and more versatile dataset than for the visual question answering task alone.

In such an architecture, the image captioning model is supposed to rely on the focus patterns that are produced by the answering model. Nevertheless, such a link between the models requires that the captioning model is indeed reacting to these externally provided focus patterns. Therefore, in this rather experimental work I tackle first the question, if this connection is actually a usable interface to interact with an image caption generator and then provide a baseline by actually connecting such models.

In Chapter~\ref{Chapter2}, I summarize the necessary background knowledge to understand the experimental setup and to further motivate my research questions. The background also contextualizes this thesis with respect to related work. In Chapter~\ref{Chapter3}, I provide a description of the spatial attention aware caption generator which is used for the experiments and how such a generator is trained. In addition, I propose three different visual focus pattern constraints on the caption generator. In Chapter~\ref{Chapter4}, I test the effect of these different visual constraints. Then in Chapter~\ref{Chapter5}, the captioning network is connected with the focus patterns of a visual question answering model. I state the results of my experiments and discuss whether these provide considerable insights based on the following research questions:

\begin{enumerate}
	\item Is the asserted control on a caption generator's attention causing the output to be different from what would have been otherwise produced?
	\item Are spatial attention forcing methods effective mechanisms to control the caption generation process in a predictable way?
	\item Is externally produced spatial attention an useful interface for image captioning models to integrate their capabilities into tasks like VQA?
\end{enumerate}

Finally, I summarize my work in Chapter~\ref{Chapter6} and propose further directions.

\clearpage


\chapter{Backgrounds} 

\label{Chapter2} 


\section{Recurrent Neural Networks for Text}
\label{sec:rnn}

As introduced above, one component of a deep learning network for both visual question answering and image caption generation has the purpose to model the textual input for the task. In this chapter I give a short overview of the necessary terms to understand my experimental setup. Therefore I briefly describe how written language is represented for a deep learning task and which network architectures are generally used to model language. Finally, I will introduce the concept of focusing on words for machine translation and how this is helpful, but different from the proposed procedures in this work.

\subsection{Representing Sentences}

When we train a language model, then we have to prepare the textual dataset to be processable by the computer. For this purpose every word is replaced by a unique number that we call the word's \textit{encoding}. These terms are from now one used interchangeably. For example, the sentence 
\begin{verbatim}
	<S> A dog is laying near a bike <E>
\end{verbatim}

can be encoded with the following sequence of numbers 

 \begin{verbatim}
 1 3 4 5 6 7 3 8 2 0 0 0 0 0 0 0
 \end{verbatim}

\clearpage
 
As we can see, the words are usually handled case-insensitive, so that both ``A'' and ``a'' become 3. Furthermore, usually a synthetic start word \textit{<S>} respectively an end word \textit{<E>} is introduced, which is here mapped to 1 or 2. Since we can have sequences of different length, we also add a \textit{NULL} or \textit{PAD} word to achieve same lengths. This word is here mapped to 0.

A language model's ability to tell something is constrained to its set of known words. This set of words is what we call a \textit{vocabulary} which is derived from large textual corpora. Therefore a vocabulary could become very large, but for reasons of performance, it is are normally restricted to a specific amount of words. For example, a large human written text corpus might contain a significant amount of individual spelling errors. The resulting erroneous words are not useful for the language learning task. Thus we could simply ignore them by specifying a small threshold for word occurrences. Given this constraint, only words that occur more often than the threshold get into the vocabulary. Along with this, all the ignored words are handled as \textit{unknown}. If the remaining corpus is large enough, training sentences with unknown words are discarded from the dataset. The handling of unknown words in language models \parencite{gulcehre_pointing_2016,sennrich_neural_2016} is an own field of research, which I don't describe in more detail here. 


\subsection{Modelling Sentences Using Words}
\label{sec:modelling_sentences}

When modeling language we usually want to predict the next word given its previous words. A traditional approach to achieve this is the \textit{n-gram} model where the word prediction only relies on its previous \textit{n} words \parencite{goodman_bit_2001}. The model is trained by simply determining all n-tuple counts within a corpus. Then for example, given a randomly chosen start word, the next probable word is naively determined by looking at all tuples that contain the $n-1$ previous words and choose the last word of the tuple with the most occurrences. Using n-grams as language models is strongly constrained by the size of the available training corpus. All co-occurrences that are not part of the corpus cannot be estimated and as such, there are soon word sequences under the language model that have zero probability. 

\clearpage

A successful technique to overcome this limitation had been introduced by \citet{bengio_neural_2003} who added another level of complexity. In contrast to the simple word encodings as mentioned above, they represented words in terms of high-dimensional \textit{features vectors} \parencite{bengio_neural_2003}. These rather semantic representation of words, nowadays called \textit{word embeddings}, allow a language model to rely on the similarity between words. Humans intuitively see that words can be grouped in a meaningful way, for example that \textit{dog} and \textit{cat} are more related to each other than \textit{dog} and \textit{bank account}. Given this world knowledge we would easily guess that the sentence \textit{The cat is eating the food} is more probable than \textit{The bank account is eating the food}, when we have seen \textit{The dog is eating the food} before. In analogy to this, language models do better, when words are represented by their semantic meaning, because then language models are able to apply word probabilities also based on similarity \parencite{mikolov_efficient_2013,pennington_glove:_2014}. 

Appropriate candidates to handle the additional complexity introduced with word embeddings are neural networks. \citet{schwenk_training_2005} used successfully a neural network as a language model on transcripts for speech recognition, but they were restricted to a predefined window of previous words. Later, \citet{mikolov_recurrent_2010} suggested recurrent neural networks to overcome this limitation and to model sentences by using all previous words. 

From an architectural view-point a recurrent neural network \parencite{elman_1990} is similar to a simple deep feed-forward network with three layers: an input layer, a state layer and an output layer. Such a three-layer neural network is called recurrent, because the network's state depends on the previous one and the learned parameters are shared among each layer during application on the input sequence. In mathematical terms we describe this network with

\begin{align}
	i^{(t)} =& \textbf{U} x^{(t)} \\
	h^{(t)} =& \text{tanh}(\textbf{W} h^{(t-1)} + i^{(t)} + b_{h}) \\
	o^{(t)} =& \textbf{V} h^{(t)} + b_{o}
\end{align}

\clearpage

where $b_{h}$ and $b_{o}$ are learned bias parameters and the weights matrices \textbf{U}, \textbf{W} and \textbf{V} are shared among the input, state and output layer at each time-step \textit{t}.

A recurrent neural network as a language model is now applied on words in sentences. Therefore, given an initial state, the network's state is adjusted at each word using the previous state. As a consequence, the state at each time-step is also based on all previous words in a sentence. As a language model, the input layer is usually also mapping the words to the according feature vectors. These word embeddings can be pre-trained ones or learned during training on-the-fly. In the end, the results of the output layer must be converted back to words of the vocabulary. A simple approach is to apply the softmax function on the output to model a probability distribution over the whole vocabulary at the time-step.

\begin{align}
y^{(t)} =& \text{softmax}(o^{(t)})
\end{align}

Then the word with the highest probability in the vocabulary is chosen as the predicted next word.\footnote{Using the highest probable word should result in same sentences for same start signals. This deterministic behavior is interesting for evaluation of manipulations to such a network, because when we manipulate the internal state, then the sentences should indeed vary.} Therefore, when training such a network, we use the final output layer results at each time-step and try to push them towards a probability distribution that is putting the most mass on the correct word in a sample by minimizing the negative log-likelihood

\begin{align}
	L^{(t)} =& -\log(p(y^{(t)}|x))
\end{align}

which is equal to minimizing the cross-entropy between the distributions. This loss is given for a specific time-step, although the trainable parameters are shared across all of them and the state is depending on the previous one. Therefore the gradient computation relies on back-propagation through time \parencite{domany_btt:_1995}. 

\clearpage

We can now use the language model after training to generate a sentence by determining an initial state and feeding a start word. Then the predicted next word becomes the input for the next time-step. The network feeds itself until the end word is produced. A similar technique called \textit{teacher forcing} is used to improve the training. Here, the network is feed the ground-truth word at each time-step instead, so that it receives a training hint and is not deviating too much from the ground truth sequence. On the downside, this might reduce generalization of the network as discussed by  \citet{bengio_scheduled_2015}.

A major problem, which often occurs during training of recurrent networks, are exploding or vanishing gradients. Here, the network is hindered to learn something useful from sequences with long-term dependencies, because the resulting training signal gets either too small or too strong \parencite{bengio_learning_1994}. To overcome this problem,  \citet{hochreiter_long_1997} introduced an effective gating mechanism. 


These \textit{Long Short-Term Memory} (LSTM) networks are learning to keep and forget training signals at certain time-steps \parencite{gers_learning_2000}. To achieve this, the network architecture of a normal recurrent network is enhanced with an input, forget and output gate around the state computation as shown respectively in the equations (2.6-8).

\begin{align}
g_{i}^{(t)} =& \sigma(\textbf{W}_{i} h^{(t-1)} + \textbf{U}_{i} x^{(t)} + b_{i}) \\
g_{f}^{(t)} =& \sigma(\textbf{W}_{f} h^{(t-1)} + \textbf{U}_{f} x^{(t)} + b_{f}) \\
g_{o}^{(t)} =& \sigma(\textbf{W}_{o} h^{(t-1)} + \textbf{U}_{o} x^{(t)} + b_{o}) 
\end{align}

Given these additional gate computations, the network is enabled to learn, whether a signal is useful for the task and thus can be kept or if the signal can be withdrawn by applying the results of the according sigmoid activation using the Hadamard product as shown in the equations (2.9-11). 

\clearpage

\begin{align}
r^{(t)} =& \text{tanh}(\textbf{W}_{r} h^{(t-1)} + \textbf{U}_{r} x^{(t)} + b_{r}) \\
c^{(t)} =& g_{f}^{(t)} \odot c^{(t-1)} + g_{i}^{(t)} \odot r^{(t)}\\
h^{(t)} =& g_{o}^{(t)} \odot \text{tanh}(c^{(t)})
\end{align}

We can see that the computation in equation (2.9) represents a normal RNN. An additional state loop is introduced in equation (2.10) as an extension. This internal LSTM loop references directly the preceding internal state $c^{(t-1)}$ and indirectly also the previous network state $h^{(t-1)}$. Here the long-short term memory network combines the previous internal state with the recurrent state computation $r^{(t)}$ by using the gates. Finally, we notice in equation (2.11) that the output gate is not applied on the network's output at the time-step, but on the new internal state. In this way the state for the succeeding time-step is adjustable. As a result, the network is enabled to effectively learn as well long-term dependencies, which are e.g. usual for the German language or primary ingredients of certain machine translation problems.



\subsection{Word Attention for Translation}
\label{sec:align_translate}

The machine translation task is intuitive and exemplary to explain the focus patterns which were introduced at the beginning. For machine translation, sentences of one language are supposed to be translated automatically into sentences of another one. For example, the English sentence \textit{A dog is laying on the street} is supposed to be translated into the German counterpart \textit{Ein Hund liegt auf der Strasse} by using an appropriate language model and a translation procedure. A simple approach would describe a statistical alignment between the words of these languages based on their position e.g. \textit{dog} and \textit{Hund} as the second word in the source sentence and translation are supposed to fulfill the same semantic or syntactic function. This approach becomes problematic, when translations are of variable lengths or translated words refer to totally different positions in the translation, which can lead to very complex alignment situations. 

\clearpage

For example, the English sentence \textit{The dog was laying on the street} is supposed to be translated into the German counterpart \textit{Der Hund hat auf der Strasse gelegen}. Here, the English temporal form \textit{was laying} has been split into the German form \textit{hat} and \textit{gelegen}, which has a distant of three words in between. An alignment model would now have to memorize \textit{laying} as an internal state representation until the end of the sentence.

There are network architectures that tackle the variable length problem by decoupling the process of encoding the source sentence and generating the translation \parencite{cho_learning_2014,sutskever_sequence_2014}. In such an architecture, an encoder transforms the source sentence into a context vector. After that, the decoder produces the translation sequence while conditioned by the context context. As a result, these encoder-decoder networks are able to translate sequences of variable length by introducing a shared bottleneck. 

This computational bottleneck leads to potential loss of information. Thus, in modern neural machine translation, the recurrent networks are learning the alignments and translations jointly by repeatedly feeding the whole source sentence into the network during the generation process \parencite{luong_effective_2015}. To achieve this, \citet{bahdanau_neural_2014} proposed a context vectors that is computed for a sequence \textit{T} like the following 

\begin{align}
	c_{i} =& \sum_{j=1}^{T_{x}}{\alpha_{ij}h_{j}}
\end{align}

where the annotations $h_{j}$ are the hidden states of the encoder at each input word and the $\alpha$ weights are scores computed by an alignment model. The alignment model is supposed to predict how important the $j$-th annotation is to produce the $i$-th target word. This predicted expectation about the importance of a certain input to produce a specific output is what we call \textit{attention}. In effect, the neural network operates on the whole input sequence instead of just a single word. This incorporation of the whole input by focusing on specific parts relaxes the computational bottleneck. 

\clearpage

\begin{figure}
	\centering
	\includegraphics{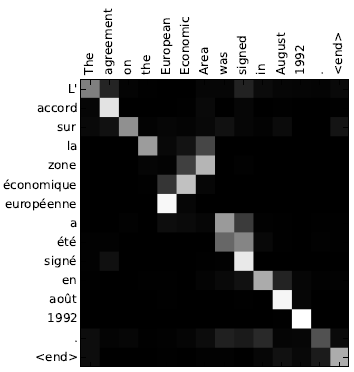}
	\caption[Example word alignment scores for a translation]{The alignment scores for a translation from English to French by \citet{bahdanau_neural_2014}. The pixel brightness indicates the weighted importance of the source words to produce a target word, which for examples shifts on \textit{européenne}, because in French adjectives come after the noun.}
	\label{fig:bahdanau_alignment}
\end{figure}

In addition, these alignment scores provide an intuitive access to the internal working of the generator for introspection as shown in Figure~\ref{fig:bahdanau_alignment}. The word attention mechanism introduced here is similar, but also different for image captioning models. The concept is similar in a sense that a captioning model tries to transform an image into a short text like a translation model tries to translate from one language into another. But it remains different, because for neural machine translation models, the attention refers to words in a source sentence, whereas for image captioning, the attention works on individual image pixels. Because of this difference, I briefly describe in the next chapter, how images are represented for visioning tasks and how vision models work internally. Then we can better understand the captioning network used in this thesis, which combines both language and vision.

\clearpage

\section{Convolutional Neural Networks for Images}
\label{sec:cnn}

A neural network for tasks like visual question answering and image caption generation has to learn a representation of the visual inputs. In this chapter I give a short summary of the necessary terms to understand my experimental setup. Therefore I briefly describe, how images are usually prepared for a deep learning task and which is the most common network architecture to model them. Finally, I will introduce the concept of spatial attention and motivate the basic assumptions for the experiments in this thesis.


\subsection{Representing Images}

\begin{figure}
	\centering
	\includegraphics[width=0.5\textwidth]{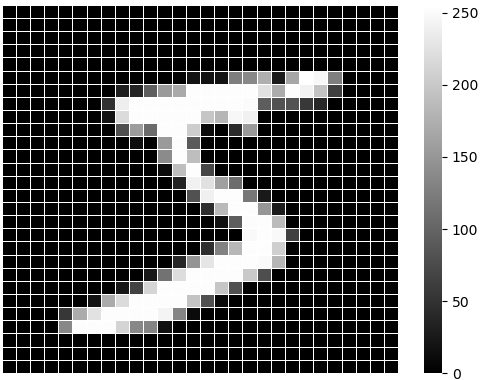}
	\caption[The number five represented as an image matrix]{A five from the MNIST database of handwritten digits \parencite{lecun_gradient-based_1998}. The number is shown as a $28\times28$ grid where each pixel is equal to a brightness value in the image matrix. The scale is given on the right and ranges from 0 to 255.}
	\label{fig:five_heatmap}
\end{figure}

Neural networks for computer vision are trained on visual input like images. For an automatic training, these images must be presented in a computer-processable way. Naturally, an image has a width $W$ and a height $H$, so that we can define an image as a matrix $\mathbb{I}^{W \times H}$ where $W$ defines the number of horizontal and $H$ the number of vertical dimensions. In this thesis, I refer to these as spatial dimensions. The values in this image matrix are describing an image at each point $p_{xy}$ with $x \in W$ and $y \in H$, which we call a \textit{pixel}.

For example, we can then represent an image in terms of the brightness level at each pixel, so that $p_{xy} \in [0, 255]$. The value range from zero to 255 is chosen, because this information can be stored in exactly one byte, which is a fundamental unit in computer science. An intuition should be given with the gray-scale image shown in Figure~\ref{fig:five_heatmap}.

\begin{figure}
	\centering
	\includegraphics[width=\textwidth]{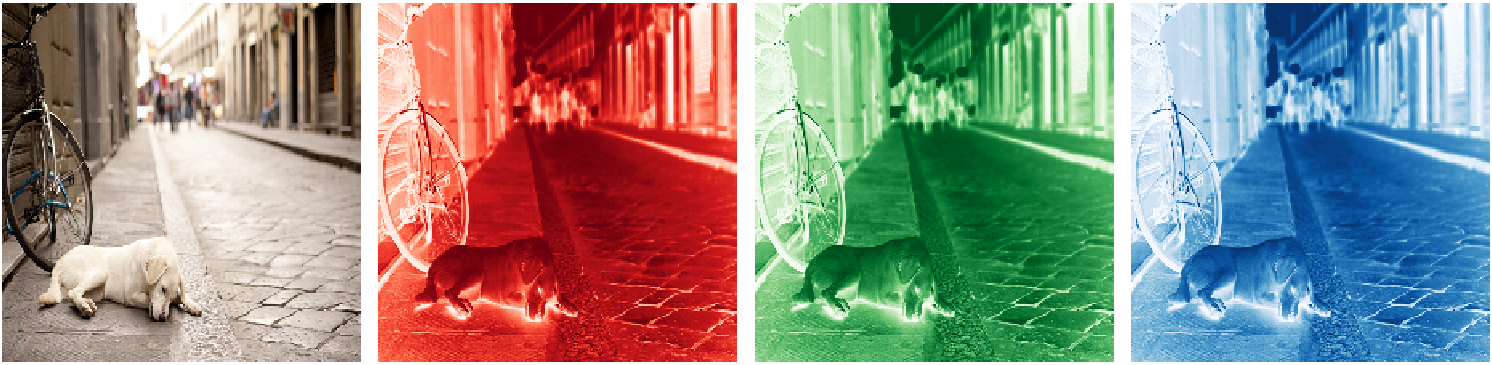}
	\caption[An image of a dog decomposed in three color channels]{An image of the MSCOCO dataset from \citet{lin_microsoft_2014}. The dog is shown as a $224\times224\times3$ image where each original pixel is also shown decomposed into its RGB channels.}
	\label{fig:dog_heatmap_rgb}
\end{figure}

Given a normally functioning visual perception, we can see that images are usually made up of colors. Using the additive color system, every color in an image can be described as the addition of the base colors \textit{red, green} and \textit{blue} (RGB). As a result, an image is additionally describable in terms of the color dimensions $C$, which we call \textit{channels}. Given this, colorful images can be  formulated as a three-dimensional matrix $\mathbb{I}^{W \times H \times C}$, where each channel also defines the according base color value at a certain pixel.\footnote{There are usually only three channels, one for each of the base colors. Sometimes, there is an additional fourth channel to describe the alpha level or transparency of a pixel.} Following this, an image has a depth $c$ where each pixel $p_{xy}$ is defined as a tuple  $(p_{xy})^{c} \in [0,255]$ with $c \in \{r,g,b\}$. An image decomposed in its color channels is shown in Figure~\ref{fig:dog_heatmap_rgb}. The three color channels add up to the original image on the left.


\subsection{Image Feature Extraction}

Given the representation of an image as a three dimensional matrix, we can now feed the image to a neural network where each of the pixel values is considered a \textit{feature} of the image. For example, when we want to model images using a normal feed-forward network, then the image matrices are flattened to $d$-dimensional feature vectors with $d = W \cdot H \cdot C$. This feature vector resembles the input layer of the network. 
We can directly see that the dimensionality of the feature vectors might become very large for already small colored images e.g. the quadratic image shown in Figure~\ref{fig:dog_heatmap_rgb} with 224 pixels is represented by a feature vector of $224 \cdot 224 \cdot 3 = 150,528$ dimensions. When we further assume that the first layer of the neural network has an appropriate size of 1,000 hidden units to handle the input complexity, then there are already about 150 million parameters to be learned. 

A neuron in such a network is operating on the whole input, but far distance spatial relations between pixels are unlikely. More likely are locally restricted accumulations of pixels that encode redundant, but consistent structures e.g. orientation of edges. \citet{hubel_receptive_1962} made experiments with the visual cortex of cats and proposed that there are simple and complex cells. The simple cell activation correlated directly with the spatial location of basic stimuli e.g. edges, whereas the complex cells also fired wherever the stimuli was placed, but only when the edge had a specific orientation. They assumed a hierarchical structure, where the complex cells interact with an image on an higher level of an hierarchy, whereas the simple cells provide basic input.

Given such a hierarchy, a neural network to model vision \parencite{fukushima_neocognitron:_1980} overcomes the parameter explosion by using \textit{local connections} meaning that a neuron is interacting only with a subset of the input. A modern description and widely known successful application of a convolutional neural network was already published by \citet{lecun_gradient-based_1998} for recognition of handwritten numbers and text. A small set of parameters $K$ is directly applied on the grid-like input structure $I$ using the \textit{convolution operation}. Given for example the one-channel image in Figure~\ref{fig:five_heatmap}, the convolution operation is describable as the cross-correlation following \citet{Goodfellow-et-al-2016}

\begin{equation}
	G(x,y) = \sum_{m}{\sum_{n}{I(m+x,n+y)K(m,n)}}
\end{equation}

with $K$ referred as the \textit{kernel} of a convolutional layer, $I$ as its input, $G$ as the layers resulting \textit{feature map} and $x,y$ as the spatial positions in the feature map. 

\clearpage

After each computation, the kernels position is adjusted in either horizontal or vertical  direction using an appropriate step-size while the parameters are the same.\footnote{The step size is what we call the stride of the convolution. The \textit{stride} is usually higher than 1 to save computations, because directly neighboring pixel-groups are likely to contain redundant information. Nevertheless, the actual hyperparameters for a well performing convolutional neural network are not relevant for this work.} For example, on our sample image a $5\times5\times3$ kernel with 75 parameters and stride 1 is applied $224 - 5 + 1 = 220$ times in the horizontal and vertical direction. Here, the result of each kernel computation becomes a value in the resulting feature map, which has then $220\times220\times1$ dimensions. 

The amount of simultaneously applied kernels determine the input depth for the upcoming layer. Usually, many kernels are applied at the same time, so that we can expect for our sample image an appropriate amount of 512 kernels in the first convolutional layer, so that there are $512\cdot75=38,400$ parameter which is a reduction by a factor of 3,920 in comparison to using a dense neural network.

Through \textit{parameter sharing}, the kernels are operating on the input data like a parameterized function $g(x)$, which is in scope of the convolution operator \textit{equivariant} to the input function $f(x)$ meaning that 

\begin{equation}
	f(g(x)) = g(f(x))
\end{equation}
 
Thus the kernel is directly reflecting changes while being repeatedly applied over the input at different positions. As a result, the kernel can be seen as a pattern matcher and the resulting feature maps as the kernel's detected image features.

Given the resulting feature maps after each convolution, the detected images features are propagated in a hierarchy, so that the first layers in a convolutional network are detecting local basic patterns like the orientation of edges. The upcoming layers are enabled to work on a larger receptive field, because they have the input of the sparsely connected lower neurons. The uppermost layers are then capable to detect more abstract concepts like different parts of animals which involve larger areas or the whole image. 

\clearpage

\begin{figure}
	\centering
	\includegraphics[width=\textwidth]{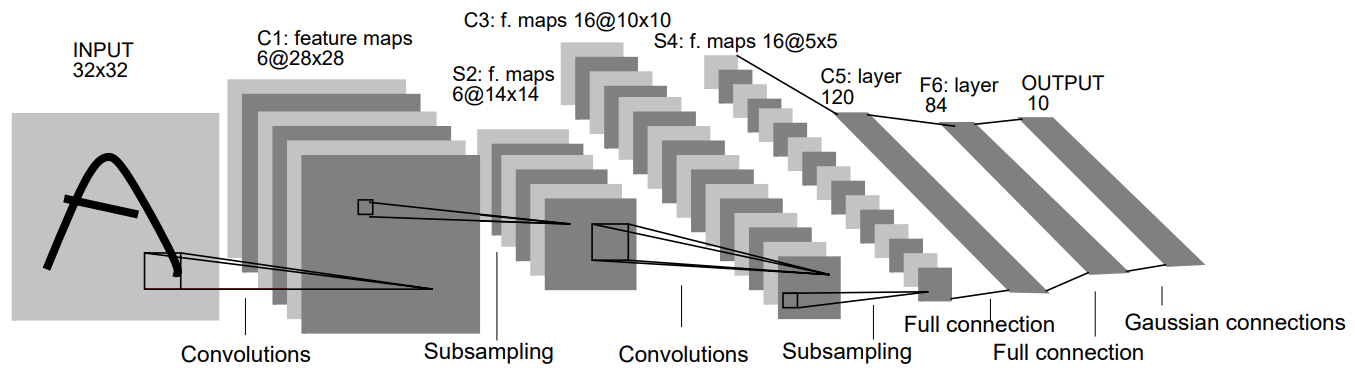}
	\caption[Architecture overview of the convolutional network LeNet-5]{Architectural view on LeNet-5 from \parencite{lecun_gradient-based_1998} for digit recognition. The networks takes images of size $32\times32\times1$. Then two convolutional blocks with max-pooling (subsampling) follow. Finally, two fully connected layers with $120$ and $84$ units are applied to classify the ten possible digits.}
	\label{fig:lenet}
\end{figure}

When the detection of certain features in an image is important, but not its exact position, then \textit{pooling} allows a convolutional neural network to become invariant to small translations of the input. 

The pooling operation provides a statistical summary of the input neurons, where for example the max-pooling operator only returns the highest value for a subset of neurons. Thus, every neuron in the group can be specialized to certain translations of an object in the input image e.g. higher layer neurons to the rotation or scaling of a dog. Then the dog might be presented in slightly changed rotation, but the network is still able to detect a dog in the image. 

A pooling layer is usually applied after one or more convolutional layers and such a group of layers is called a \textit{convolutional block}. After the last convolutional block the actual classification network is put. This network usually consists of one ore more fully connected layers and a final softmax output layer to model the distribution over a set of classes given an input image. A fully architectural view on a convolutional network is shown in Figure~\ref{fig:lenet}.

\subsection{Image Features as Object Indicators}

\citet{krizhevsky_imagenet_2012} had initially shown that a CNN with a large enough capacity is able to classify up to 1,000 different object categories. Their AlexNet consists of five convolutional blocks and a large classification network with three fully connected layers. In the ILSVRC-2012 challenge \parencite{ILSVRC15} with millions of training samples, the AlexNet had shown state-of-the-art results. The results outperformed other image classification techniques that relied on engineered pattern matchers like scale-invariant features transforms (SIFT) \parencite{lowe_object_1999} and histograms of oriented gradients (HOG) \parencite{dalal_histograms_2005}. 

Afterwards, convolutional neural networks have become the standard for image classification and object detection tasks. Nowadays, the most popular architectures are trained for thousand of categories on millions of images of the hierarchically organized ImageNet dataset \parencite{deng_imagenet:_2009}. Given the large amount of data and the high capacity, the networks learn during weeks of training useful feature representations. After training, these models are made available for the public. 

One of the most popular pre-trained computer vision models, is the very deep convolutional neural network (VGG) because of its powerful capacity and simple structure. The network processes image inputs of $224\times224\times3$ dimensions. After the input layer, the network is organised in five blocks of convolutional layers each followed by a max pooling layer. The last block consists of three large fully connected layers and a softmax layer with 1,000 classes \parencite{simonyan_very_2014}. 

To reveal the inner workings of these networks, \citet{fleet_visualizing_2014} asked the question, which input image would activate a classification neuron the most. Thus, they introduced an approximate inverse of the convolution operation, called deconvolution or transpose convolution, which takes an activation and computes a possible input. Here, a single class neuron of a trained network is maximally activated and then the backpropagation is performed down to the input image. Later, \citet{yosinski_understanding_2015} improved these visualizations with better regularization techniques. They found that the lower convolutional layers already provide a rich palette of detectors for potential higher level classification layers. On the downside, they also found in a related work that convolutional networks are easily fooled by minor changes in the input images \parencite{nguyen_deep_2015}.

\clearpage

\citet{lin_network_2013} found that global average pooling after the last convolutional block acts as a structural regularizer which helps to reduce overfitting for an overall structure. In average pooling the average values for a spatial region are computed, so that the network is less influenced by extreme values. In fact, \citet{zhou_learning_2016} found that average pooling is highly useful for localization and introspection. They assumed that such a global pooling encourages the network to learn whole object shapes instead of specific parts by forcing the correspondence between feature maps and categories. As a result they introduce class activation maps. \citet{selvaraju_grad-cam:_2017} extended this technique to a gradient based method which is useful to find the regions in an image that activates a specific object classifier the most. 

\begin{figure}
	\centering
	\includegraphics[width=\textwidth]{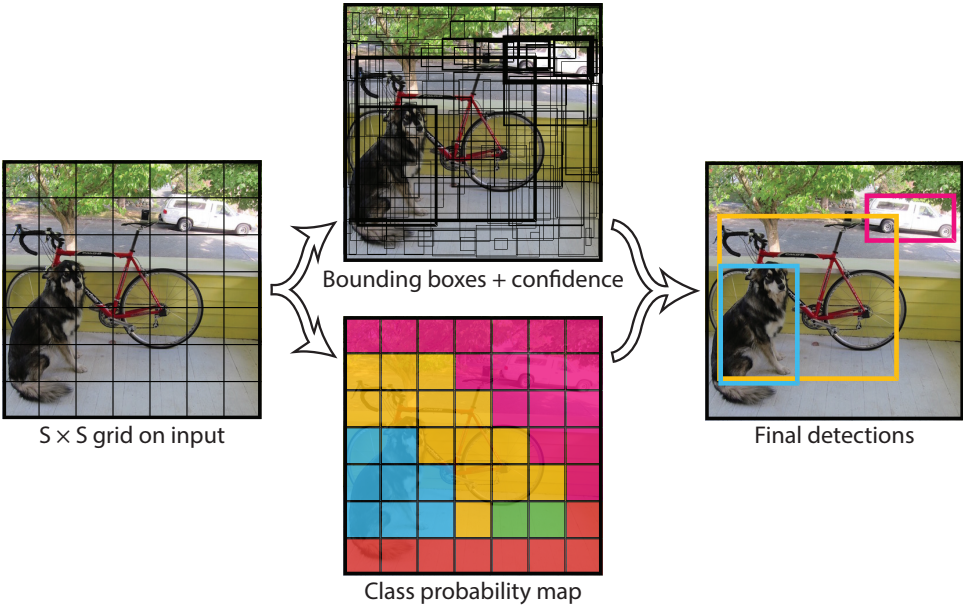}
	\caption[The conceptual idea of ``You Only Look Once'']{The conceptual idea from \citet{redmon_you_2016}. The input is decomposed in multiple boxes using a grid-like arrangement. Then the network proposes bounding boxes and computes class probabilities for each box in parallel. Finally, the bounding boxes are chosen which fit best the annotated boxes. }
	\label{fig:yolo}
\end{figure}

These introspective papers indicate that we can interpret outputs of upper convolutional layers already as a summarized spatial attention with respect to an object activation. Given this, the network is enabled to conclude the presents of certain objects. Thus, this intrinsic spatial attention leads to useful application in object detection.
\citet{ren_faster_2017} introduced the Faster R-CNN which uses the extracted image features to feed a region proposal network. This network serves as an attention guidance over the image. As such, these region proposals are input for an additional neural network, which finally performs both classification and bounding box regression. The region proposal network is trained end-to-end with the features extractor to distinguish actual object categories from the background. \citet{he_mask_2017} introduced Mask R-CNN as an extension to this framework, which produces in addition a segmentation mask for objects in the input image. In analogy, \citet{redmon_you_2016} used the ability to classify even small regions in an image for real-time object detection as shown in Figure~\ref{fig:yolo}.



These are successful examples for the capabilities of convolutional neural networks to model and localize objects in images. We have seen that outputs of the upper convolutional blocks are interpretable as spatial attention for an object in the image. In a similar way, the attention for language models is related to the produced word. Now, attentive image captioning models make use of both architectural capabilities by assuming a connection between the word to be produced and the spatial attention on an object in the images.


\clearpage

\clearpage

\section{Image Captioning Networks}

Based on the described capabilities of recurrent networks from section \ref{sec:rnn} and convolutional network from section \ref{sec:cnn}, the natural language processing tasks have evolved towards image captioning. An image captioning network is supposed to generate a short description about a presented image. 

The first emerged approaches to tackle this problem are called \textit{bottom-up} techniques. At the bottom, the convolutional neural networks are used to predict possible words or objects categories for an image by using the classification capabilities. Then these word predictions are taken to fill sentence templates  \parencite{elliott_image_2013,fang_captions_2015,kulkarni_baby_2011,kuznetsova_collective_2012}. The architectural separation of word prediction and sentence generation makes end-to-end training difficult.

The modern \textit{top-down} approaches combine the vision and language networks to learn a multi-model representation for both extracted image and learned sentence features, which allows an end-to-end training \parencite{chen_minds_2015,donahue_long-term_2017,mao_deep_2014,vinyals_show_2015}. 

Nowadays, these top-down architectures are also often involved in visual question answering tasks which makes them an interesting method for this thesis. In this section, I will present such a network architecture proposed by \citet{karpathy_deep_2017}. In addition, image caption networks understood as multi-modal models allow an intuitive implementation of attention mechanisms. I will review the use of spatial attention for image captioning as described by \citet{xu_show_2015} and based on this, motivate my basic working assumptions for the experiments in this thesis.

\subsection{Modelling Images and Words Together}

\citet{frome_devise:_2013} introduced a deep visual-semantic embedding model to learn vision and language together. Given such a trained model, they have shown to make successful predictions also about unseen data, for example to retrieve images given a text or to find texts given an image. 

\clearpage

\begin{figure}
	\centering
	\includegraphics[width=0.7\textwidth]{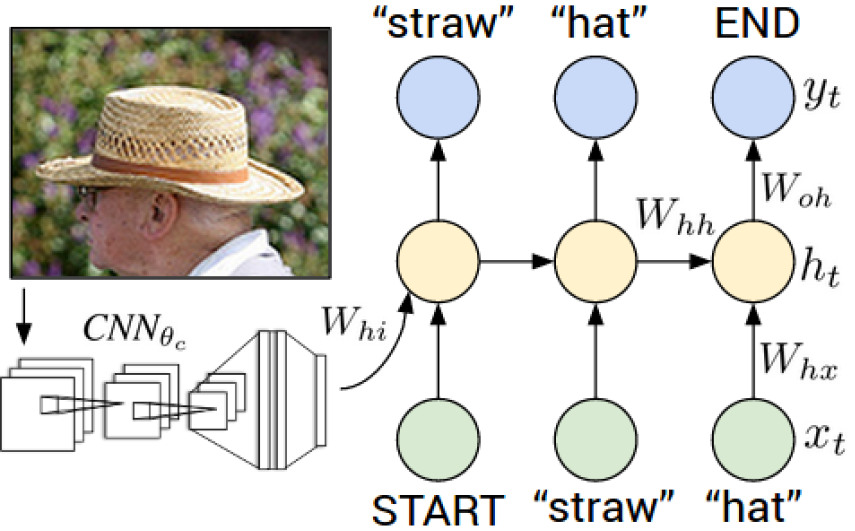}
	\caption[Architectural view on a multimodal RNN for captioning]{The architectural view on the multimodal RNN from \citet{karpathy_deep_2017}. A CNN is used to extract image features. The recurrent network is conditioned in the first time-step on these feature and the start word. Inputs are shown in green, network states in yellow and outputs in blue.}
	\label{fig:karpathy}
\end{figure}

Inspired by this work, \citet{karpathy_deep_2017} use a multi-modal model to produce  descriptions of specific image regions. They propose a variant of a bi-directional recurrent network \parencite{schuster_bidirectional_1997}. 

As shown in Figure~\ref{fig:karpathy}, the recurrent network is receiving at the initial time-step the visual features extracted by a convolutional network and a start word. Then the conditioned recurrent network predicts at each time-step a word given the context from the previous time-steps. More formally, \citet{karpathy_deep_2017} describe the model with the following equations 

\begin{align}
	b_{v} =& \textbf{W}_{hi}[\text{CNN}_{\theta_{c}}(\text{I})] \\
	h_{t} =& f(\textbf{W}_{hx}x_{t} + \textbf{W}_{hh}h_{t-1} + b_{h} + \underline{\mathbb{1}(t=1) \odot b_{v}}) \\
	y_{t} =& \text{softmax}(\textbf{W}_{oh} h_{t}+b_{o})
\end{align}

where $\textbf{W}_{hi}$, $\textbf{W}_{hx}$, $\textbf{W}_{hh}$, $b_{h}$ and $b_{o}$ are trainable parameters and $\text{CNN}_{\theta_{c}}(\text{I})$ are the extracted image features of a pre-trained R-CNN \parencite{girshick_rich_2014}. 

\clearpage

We can see, that this formulation is similar to the normal recurrent neural network described in section \ref{sec:modelling_sentences}, but extended to the add operation for the image features at the first time-step. Apart from this, \citet{karpathy_deep_2017} found that using a bi-directional RNN further improves the model capabilities, because the network is better on longer contexts. This founding aligns with the machine translation model from \citet{bahdanau_neural_2014} who used a bi-directional neural network with attention to involve larger contexts.

\subsection{Spatial Attention for Word Prediction}
\label{sec:align_visual_and_language}

In section \ref{sec:align_translate}, I introduce a soft alignment model that computes scores to produce a context vector by a weighted set of annotations. These scores are interpretable as a word attention over the whole source sentence. \citet{xu_show_2015} transferred the soft alignment idea to the image captioning task and approximated a context vector $\hat{\textbf{z}}_{t}$ for each time-step by computing the weighted sum over all image annotations

\begin{equation}
	\mathbb{E}_{p(s_{t}|\text{a})}[\hat{\textbf{z}}_{t}] = \sum_{i=1}^{L}(\alpha_{ti} \textbf{a}_{i})
\end{equation}

where the annotations $\textbf{a}_{i}$ are now $D$-dimensional pixel-wise image features extracted by a pre-trained convolutional neural network and $p(s_{t}|\text{a})$ is the probability of an attention location random variable $s_{t}$ given the $L$ extracted image features

\begin{equation}
	\text{a} = \{\textbf{a}_{1},...,\textbf{a}_{L} \}, \textbf{a}_{i} \in \mathbb{R}^{D}
\end{equation}

I have to notice here, that the pixel-wise image features contain one piece of information about each features map extracted for a specific receptive field. Therefore, this ``in-depth'' slicing along the feature dimensions retains the spatial information of the input image in a compressed form. 

\clearpage

Furthermore, \citet{xu_show_2015} concluded from \citet{baldi_dropout_2014} that the expectation over the context vector  $\mathbb{E}_{p(s_{t}|\text{a})}[\hat{\textbf{z}}_{t}]$ can be approximated as the marginal likelihood over the attention location by using a single-forward pass through the \text{softmax} function. Therefore, \citet{xu_show_2015} showed that an attention model can be described as capable to approximate the likelihood over the attention locations with the following equations 

\begin{align}
	e_{ti} =& f_{\text{att}}(\textbf{a}_{i}, h_{t-1}) \\
	\alpha_{ti} =& \frac{\exp(e_{ti})}{\sum_{k=1}^{L}\exp(e_{tk})}
\end{align}

when the previous hidden state activation $h_{t-1}$ is interpreted as the linear projection of the context vector $\hat{\textbf{z}}_{t}$. They called this methodology \textit{deterministic soft attention}, because the resulting algorithm is fully differentiable, so that stochastic gradient descent methods with back-propagation are applicable.

\begin{figure}
	\centering
	\includegraphics[width=0.8\textwidth]{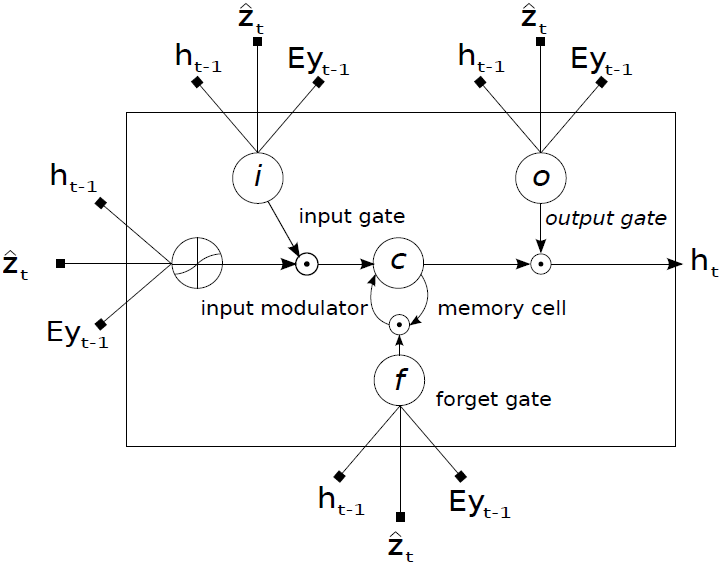}
	\caption[An LSTM cell extended for spatial attention]{The extended LSTM cell from \citet{xu_show_2015}. At the first glance, a normal LSTM architecture is shown. However, the inputs are extended to include also the previous word embedding and the current context vector. The current context vector is derived from the previous state and the current spatial attention over the image. In effect, there are multiple inputs, but there is only the current state as an output.}
	\label{fig:xu_lstm}
\end{figure}

\citet{xu_show_2015} extended a normal LSTM to incorporate the visual context vector and the previous predicted word as shown in Figure~\ref{fig:xu_lstm}. The state output is then used in combination with the visual context and the previous word embedding to model an output distribution over the whole dictionary. Following this, the context vector and the previous word are applied twice, once during encoding for the hidden state computation and once at decoding for the next word prediction.

The spatial soft attention methods for image captioning has inspired a lot of other works. For example, \citet{you_image_2016} extended the framework by adding visual attribute detectors for each word in the vocabulary. Then they used an attention mechanism to rank the visual attributes to induce them as semantic features into the image captioning task. \citet{anderson_bottom-up_2017} combined the top-down and bottom-up approaches by using a Faster R-CNN to propose image regions as feature vectors and then weights these proposals using a soft attention mechanism. 

\clearpage

\clearpage

In contrast to these related works, I examine the direct interaction with the soft attention mechanism. Therefore, I train such a spatial attention aware network and try to assert control over the model after training by using this connection to evaluate its usefulness as an external interface for other tasks.

\subsection{Controlling the Spatial Attention}
\label{sec:controlling_attention}

So far, there had been several attempts to achieve more control over neural language generation. \citet{anderson_guided_2017} controlled the output process of a captioning model at test time with an enhanced beam search. Here, an external system generates image tags as a control signal at the decoder level. They showed that involving the generated hints during the generation process actually improved the performance for out-of-domain captioning. Although this approach worked, there is no attention effecting mechanism involved. \citet{zarriess:inlg18} evaluated a ``trainable decoding'' approach that inserts task-specific concerns into the decoding process.

However, directly interfering with the spatial attention mechanism after training has been tried less often. \citet{cornia_show_2018} trained a captioning model not only to learn the distribution for images and sentences, but also for bounding boxes and noun chunks. In addition, the model had to learn when to switch between boxes. As a result, the captioning model was controllable by a bounding box sequence provided as an input to the network at test time. Although this approach had been shown to work well, they explicitly designed the model to be controllable.

In contrast to the previous approaches, I assume that a captioning model with spatial soft attention is inherently controllable. In such a sense, my approach is an inverse of the visual grounding task. \citet{rohrbach_grounding_2016} localized phrases within an image by deriving bounding boxes from the spatial attention of a specially trained model. I try to reverse this direction and fix the attention to manually chosen parts of the image after training to generate captions about that region. 

Given the knowledge presented in this chapter, I propose in the following two basic working assumptions on which I base my experiments in this work.

\clearpage

\begin{figure}
	\centering
	\includegraphics[width=\textwidth]{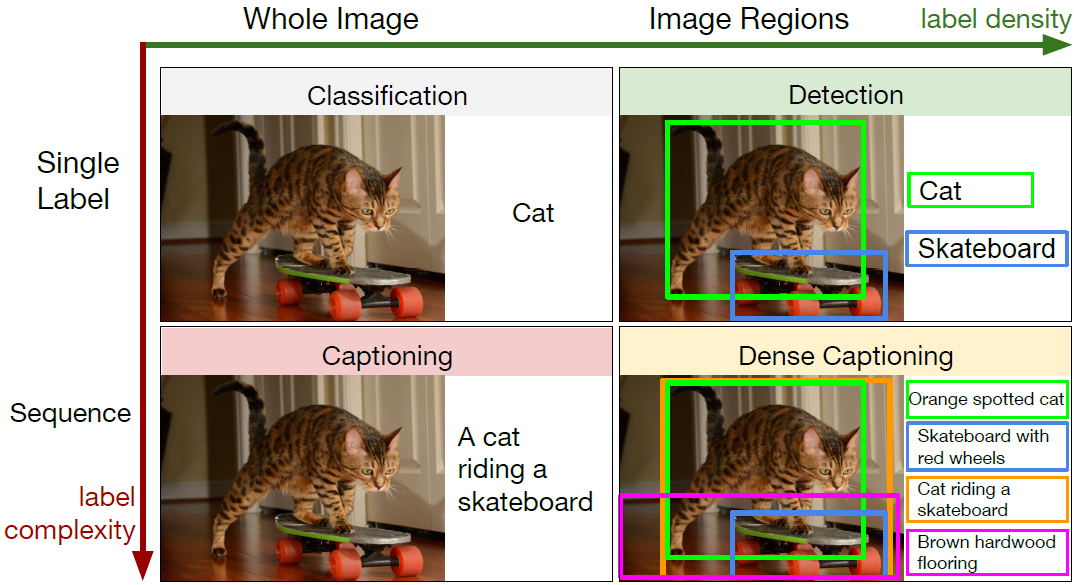}
	\caption[An overview over the dense captioning task]{The dense captioning task as proposed by \citet{johnson_densecap:_2015}. The dense captioning task combines both the generation of descriptions and the detection of individual object categories. For training, the detection task requires bounding boxes and object labels while the captioning task requires sentences. A dataset ideally provides these.}
	\label{fig:johnson_dense}
\end{figure}

\paragraph{Working Assumption 1: Inherent Spatial Awareness.} My first assumption is that a spatial  attention-aware captioning model, which is trained for the whole image captioning task, automatically includes the dense captioning task on the same domain, when the model is trained with enough samples that contain and describe individual objects in the images. 

\citet{johnson_densecap:_2015} introduced the dense captioning task as shown in Figure~\ref{fig:johnson_dense}. They proposed a Fully Convolutional Localization Network (FCLN) architecture that extends the Faster R-CNN network with bilinear interpolation \parencite{jaderberg_spatial_2015}. Thus they are allowed to train the network end-to-end without external region proposals, but they restricted themselves to the rectangular regions and pass them only towards the first time-step. \citet{xu_show_2015} already argued that deterministic soft attention is more flexible, because the network is able to incorporate the whole image in contrast to only fixed region proposals.

\clearpage

\begin{figure}
	\centering
	\includegraphics{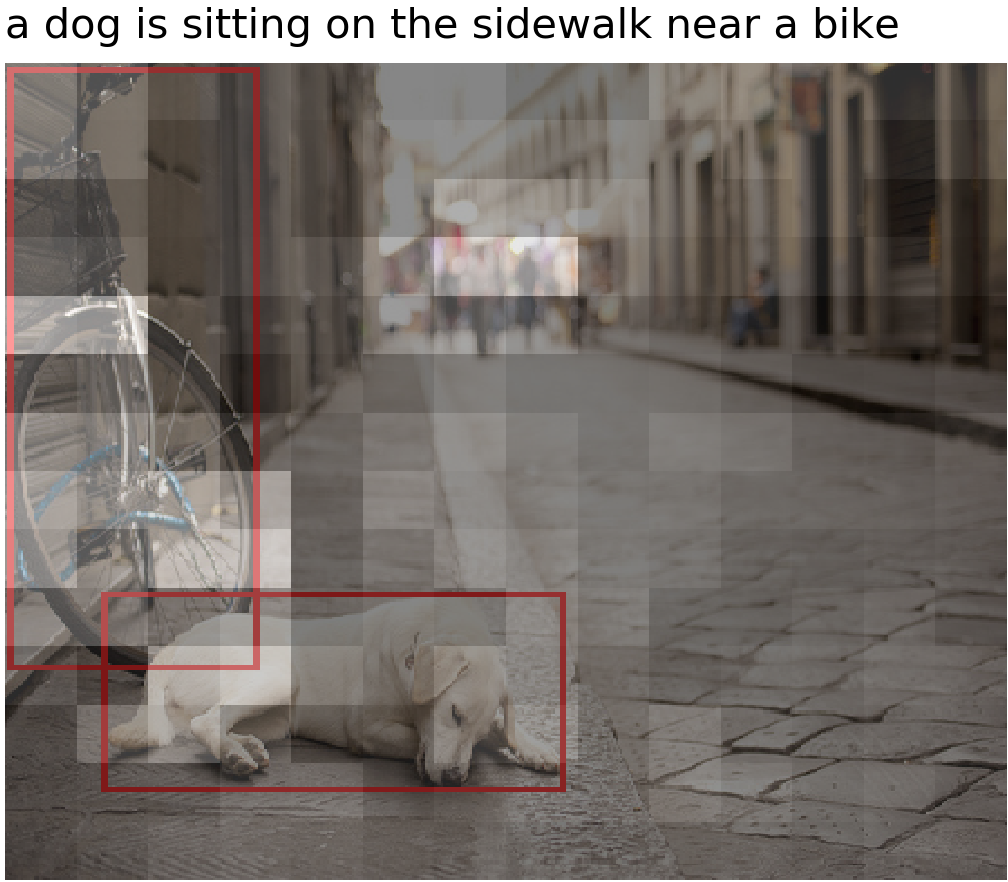}
	\caption[A produced caption for a caption generator with spatial attention]{A caption generated by an image captioning model with spatial attention. The spatial attention is pixelated and summed up over all time steps. In addition, the dog and the bicycle are framed with the corresponding bounding boxes.}
	\label{fig:dog_boxes}
\end{figure}

\paragraph{Working Assumption 2: Inherent Object Alignment.} Secondly, I intuitively expect that the generative features of the model would highly correlate with spatial attention on distinct objects in the images.  

Since its introduction, the attention involved, when generating a particular output sequence, has been viewed as providing a potentially explanatory insight in the internal workings of the model. A common observation is that spatial attention-aware image captioning models are attending to different parts of an image that we intuitively interpret as salient objects. 

For example as shown in Figure \ref{fig:dog_boxes}, when producing a caption, the model is attending most of the times to the distinct objects in the image: the bike, the dog and the group of people in the background. \citet{xu_show_2015} observed that this behavior is enforceable by learning a gating scalar $\beta$ which is then applied to the visual context vector with $\beta \cdot \hat{\textbf{z}}$. I see a possible explanation for this effect in the resulting model's ability to weight the visual context vector dynamically. This allows the model to separate visual words from textual words as also described by \citet{cornia_show_2018}.
 

\chapter{Methods} 

\label{Chapter3} 


Sequential deep learning models with an attention mechanism are able to additionally involve context by ``attending to'' select parts of the whole input sequence at each time step. This had been initially shown to be helpful as word attention for neural machine translation, which operates on sequences of words \parencite{bahdanau_neural_2014, luong_effective_2015}. Later, deep learning networks with attention were capable to jointly learn the alignment of images and language \parencite{karpathy_deep_2017, xu_show_2015}. These top-down captioning models with spatial attention  focus on different parts of the extracted image features to produce a word sequence. Here, the encoded spatial image features are either fed to the network once at the beginning or at each time step. In this chapter, I present the image captioning architecture chosen for my experiments and introduce the interface methods which I use to manipulate the model's spatial attention.

\section{The \textit{Show, Attend and Tell} Caption Generator} 
\label{sec:shatt}

The image captioning network architecture chosen for the experiments is from \citet{xu_show_2015}, who's attention mechanism is described in section \ref{sec:align_visual_and_language}. The model predicts an explicit spatial attention vector over the image for each word, which allows direct manipulation of the model's attention. 

Following \citet{xu_show_2015} I conclude that my working assumptions for the experiments are covered using their architecture:

\begin{enumerate}
	\item The spatial soft attention is supposed to include the dense captioning task as the model is capable to incorporate every region in the image for the caption generation process.
	\item The model has shown to correlate objects within the images with spatial attention, when an according gating scalar is computed.
\end{enumerate}

Last but not least, they use standard techniques and have been highly influential for further research on attention based neural networks. Therefore this thesis can be easier contextualized with other works in this field.

\subsection{Reimplementation and Modifications}
\label{sec:reimplementation}

The network implementation referenced by \cite{xu_show_2015} was not available anymore.\footnote{Only later, I found that the source code is in a repository called \textit{artic-captions} \parencite{xu_theano_2016}} Therefore, I re-implemented the network in Keras \parencite{francois_chollet_keras_2015} by following an implementation of the network from \citet{choi_2016} in TensorFlow \parencite{tensorflow2015-whitepaper}.\footnote{My source code is documented at \url{https://github.com/phisad/keras-shatt}} Additional reasons for an own implementation were an easier modification of the network architecture and a proof-of-concept, if the presented results are actually achievable given the described procedure.

\paragraph{The Image Extractor.}
The image feature extractor used by \citet{xu_show_2015} is the VGG-19 from \citet{simonyan_very_2014} which had been pre-trained on the ImageNet dataset \parencite{deng_imagenet:_2009} without fine-tuning. Thus, I used the according public model provided by the Keras framework also without fine-tuning. These model's weights are directly converted from the Oxford VGG and should therefore be the same. \citet{xu_show_2015} use the $14\times14\times512$ image features of the fourth convolutional layer in the fifth convolutional block before max-pooling. As a modification, I use the image features after max-pooling as suggested by \citet{yang_stacked_2015}. This would result into smaller features maps of $7\times7\times512$, thus I double the input size from $224\times224\times3$ to $448\times448\times3$. Experiments with both configurations have shown that the larger input size is indeed producing better scores.

\clearpage

\paragraph{The Initializer Network.}

The initial LSTM states were predicted using the mean image annotation of the resulting image feature maps. The mean vector $m_{a} \in \mathbb{R}^{D}$ was calculated per image annotation $A \in \mathbb{R}^{L \times D}$ like

\begin{equation}
	m_{a} = \frac{1}{L} \sum_{i}^{L}(A(i,j)) 
\end{equation}

with $D$ as the number of image feature maps and $L$ as the amount of extracted image features per feature map. Then the mean annotation vector was fed to the following simple feed-forward neural networks with tanh-activation

\begin{align}
	c_{0} =& \text{tanh}(\textbf{W}_{c_{0}} m_{a} + b_{c_{0}}) \\
	h_{0} =& \text{tanh}(\textbf{W}_{h_{0}} m_{a} + b_{h_{0}})
\end{align}

where $\textbf{W}_{c_{0}}  \in \mathbb{R}^{H \times D}$ and $\textbf{W}_{h_{0}} \in \mathbb{R}^{H \times D}$ were trainable parameters and $b_{c_{0}}$, $b_{h_{0}}$ the bias terms. 

\paragraph{The Attention Network.}

As part of the encoder, the attention network computed the image feature attention to derive the context vector. The attention network computed first an image feature projection which was fed into each time-step 

\begin{equation}
\text{P}_{a} =  \text{A} \times \textbf{W}_{\text{P}_{a}}
\end{equation}

learning the trainable parameters $\textbf{W}_{\text{P}_{a}} \in \mathbb{R}^{D \times D}$ for the image annotations $A \in \mathbb{R}^{L \times D}$ where $D$ is the number of image feature maps and $L$ the amount of extracted image features per feature map. Furthermore, the hidden state was projected from $H \rightarrow D$ to align the number of dimensions by computing

\begin{equation}
p_{h}^{(t)} =  \textbf{W}_{p_{h}} h^{(t-1)} + b_{p_{h}}
\end{equation}

\clearpage

with trainable parameters $\textbf{W}_{p_{h}} \in \mathbb{R}^{D \times H}$ so that $p_{h}^{(t)} \in \mathbb{R}^{D}$. This projection of the hidden state into D dimensions was acting like a bias term on the computed image feature projection, in such a sense that the previous hidden state projection was added to each feature map. Afterwards, the ReLU activation \parencite{hahnloser_permitted_2001} was applied like

\begin{equation}
	\text{P}_{a}^{(t)} = \text{ReLU}(\text{P}_{a} + p_{h}^{(t)} )
\end{equation}

so that $\text{P}_{a}^{(t)} \in \mathbb{R}^{L \times D}$ now represents the image annotations conditioned on the previous state. This allows the network to adjust specific feature map signals based on the previous state. Then the actual image attention is determined by summing up along the feature map dimension and applying the softmax function as described in section \ref{sec:align_visual_and_language}

\begin{equation}
	 \alpha^{(t)} = \text{softmax}(\sum_{j}^{D}[\text{P}_{a}^{(t)}(i,j)])
\end{equation}

so that the spatial attention is given by $\alpha \in \mathbb{R}^{L}$, $\alpha \in (0,1)$ and $\sum{\alpha} = 1$. 

\paragraph{The Encoder Network.} Given the spatial attention, the network's context vector was defined following \citet{xu_show_2015} with 

\begin{equation}
z^{(t)} = \sum_{i=1}^{L}(\alpha_{i}^{(t)} \text{A}(i,j))
\end{equation}

so that $z^{(t)} \in \mathbb{R}^{D}$ with the image features at a spatial location weighted by the predicted attention conditioned on the previous hidden state. This allowed the network to adjust the weightings of specific image regions. In addition, I computed the gating scalar $\beta$ using 

\begin{equation}
\beta^{(t)} = \sigma( \textbf{W}_{\beta} h^{(t-1)} + b_{\beta})
\end{equation}

\clearpage

with trainable parameters $\textbf{W}_{\beta} \in \mathbb{R}^{H}$ so that $\beta \in \mathbb{R}$ as described by \citet{xu_show_2015} on \textit{Doubly Stochastic Attention}. They found that given $\beta$ the model is putting more attention weights on objects in the images, when applying the gating scalar to the context vector like

\begin{equation}
\hat{z}^{(t)} = \beta^{(t)} \cdot z^{(t)} 
\end{equation}

which allows the network to distinguish between visual and textual words as also described by \citet{cornia_show_2018}. Finally, the word embedding for the previous word was computed and combined with the gated context vector as input for the LSTM at the specific time-step following 

\begin{equation}
	x^{(t)} = \textbf{E} y^{(t-1)} \oplus \hat{z}^{(t)}
\end{equation}

with the trainable word embedding matrix $\textbf{E} \in \mathbb{R}^{D \times V}$ where $D$ is the number of word dimensions, $V$ is the vocabulary size and the word is given by an one-hot encoded vector $y$ of size $V$. Then we get the input vector $x^{(t)} \in \mathbb{R}^{H}$ using the concatenation operation $\oplus$, when we assume that the word and image embedding dimension are equal, so that $D_{\text{E}} + D_{\text{A}} = 2 \cdot D = H$.

\paragraph{The Decoder Network.}
The decoder network computed the output word distribution over the vocabulary given the LSTM state computation based on the input $x^{(t)}$ and the previous state $h^{(t-1)}$ as described in section \ref{sec:modelling_sentences}. \citet{xu_show_2015} designed the decoder as a deep output network with dropout. Given their description along with \citet{choi_2016}, I implemented the following

\begin{equation}
	o^{(t)} = \textbf{E} y^{(t-1)} + \textbf{W}_{oz} \hat{z}^{(t)} + \textbf{W}_{oh} \rho(h^{(t)}) + b_{o_{h}}
\end{equation}

where $\textbf{W}_{oz} \in \mathbb{R}^{D \times D}$ and $\textbf{W}_{oh} \in \mathbb{R}^{D \times H}$ are trainable parameters for the gated context vector or resulting hidden state respectively. The word embedding is the same as above and $\rho$ is the dropout operation \parencite{srivastava_dropout:_2014}. Finally, the output word distribution over the vocabulary was computed using

\clearpage

\begin{equation}
y^{(t)} = \text{softmax}(\textbf{W}_{y} \rho(o^{(t)}))
\label{eq:shatt_output}
\end{equation}

with the trainable parameters $\textbf{W}_{y} \in \mathbb{R}^{V \times D}$. Actually, I computed $V+1$ words to allow the padding word with an encoding of zero to be predicted. The specific formulation depends on the understanding of whether the pad word is part of the vocabulary. The resulting distribution vector was then usable to minimize the cross-entropy loss or to apply a sampling algorithm to produce an actual word at the time-step.


\subsection{Training Dataset and Preparation}
\label{sec:mscoco}

After the implementation, I trained the network on the MSCOCO dataset for the \emph{Captioning Challenge 2015} \parencite{lin_microsoft_2014}. The dataset consists of $82,783/40,504/81,434$ images for training, validation and test respectively. Each image is annotated with five ground-truth captions, for example the image of the validation split showing a dog and a bicycle in Figure \ref{fig:dog_boxes} has the following captions:

\begin{enumerate}
	\setlength\itemsep{-0.5em}
	\item A picture of a dog laying on the ground.
	\item Dog snoozing by a bike on the edge of a cobblestone street
	\item The white dog lays next to the bicycle on the sidewalk.
	\item a white dog is sleeping on a street and a bicycle
	\item A puppy rests on the street next to a bicycle.
\end{enumerate}

This is the same dataset that was also used by \citet{xu_show_2015}, so that I could check the scores of my implementation against the ones reported in their work. Large differences may indicate failures in the implementation while smaller deviation my be depending on use of different hyperparameters. 

\begin{table}
	\begin{tabular}{| r | r | r | r | r |}
		\hline
		Caption & \multicolumn{2}{c|}{unlimited vocabulary} & \multicolumn{2}{c|}{limited vocabulary} \\
		Length & \# with Length & accumulated & \# with Length & accumulated \\
		\hline
		8  &   3,238 (01\%) &   3,249 (01\%) &   3,114 (01\%)  &  3,124 (01\%)\\
		9  &  68,196 (17\%) &  71,445 (18\%) &  65,489 (17\%)  & 68,613 (18\%)\\
		10 &  90,207 (23\%) & 161,652 (40\%) &  87,047 (23\%) & 155,660 (41\%)\\
		11 &  88,974 (22\%) & 250,626 (63\%) &  85,803 (22\%) & 241,463 (63\%)\\
		12 &  64,009 (16\%) & 314,635 (79\%) &  61,647 (16\%) & 303,110 (79\%)\\
		13 &  40,654 (10\%) & 355,289 (89\%) &  39,000 (10\%) & 342,110 (89\%)\\
		14 &  23,612 (06\%) & 378,901 (95\%) &  22,372 (06\%) & 364,482 (95\%)\\
		15 &  13,382 (03\%) & 392,283 (98\%) &  12,610 (03\%) & 377,092 (98\%)\\
		16 &   7,704 (02\%) & 399,987 (X0\%) &   7,208 (02\%) & 384,300 (X0\%)\\
		\hline
	\end{tabular}
	\caption[The distribution of captions in the training split given by length]{The number of captions given by length and for an unlimited and limited vocabulary. The limited vocabulary was constrained to the most common 10,000 words. For training, I discarded captions that contain uncommon words. (X0 = 100\%)}
	\label{table:captions_training}
\end{table}

\paragraph{Caption preprocessing.} \citet{xu_show_2015} stated to use basic tokenization, but were unclear on the exact procedure. I used the standard Treebank word tokenizer from the \textit{Natural Language Toolkit} (NLTK) \parencite{loper_nltk:_2002} to tokenize the 414,113 training captions. In addition, I converted single digits $[0,9]$ to their corresponding words e.g. $2 \rightarrow \text{two}$, because they might occur as text and as number in the caption. Other numbers like years were removed, because they are rare and not necessary for this captioning task. 

Given this tokenization procedure, the captions length ranged from $6$ to $51$ in the training split. Nevertheless, long captions were rare and required a larger capacity of the network for decoding. Thus, I targeted a trade-off between maximal caption length and training split size. Therefore, instead of following \citet{xu_show_2015} who used all captions by sorting them into batches of same length, I deviated from this procedure and followed \citet{choi_2016} who shuffled the samples randomly into batches and reduced the set of captions to the ones with a maximal length of 16 tokens. This resulted in a total of 399,987 captions which still represented 96,59\% of the training split. 

Building a vocabulary based on this subset resulted in $22,461$ distinct tokens, but not all of them were useful. For example there were $670$ tokens that occurred less than $10$ times and were uninterpretable ones like \textit{bby, eeg, kc} or \textit{hew}. These rare occurrences required capacity on the decoder, although they were unlikely to be useful for the task. \citet{johnson_densecap:_2015} mapped words with less than 15 occurrences to the \textit{unknown} token. This would had been $16,810$ tokens to be discarded for my subset with only $5,650$ tokens appearing at least 15 times. 

\clearpage

Therefore, I proceeded as \citet{xu_show_2015} and kept the $10,000$ most common tokens. To achieve this, I first built an auxiliary vocabulary that was constrained to the most common words. Then I encoded the words in each caption and discarded those captions that contained an \textit{unknown} encoding. After this, there were $15,687$ captions less than before in the training split as shown in Table \ref{table:captions_training} and $6,545$ captions fewer in the validation split. 

In addition, I removed from all captions the dot, comma, single quote, double quotes, hyphen, parenthesis and replace \& with ''and''. The final vocabulary built from the filtered captions was reduced to $9,993$ tokens including the start and end token. Furthermore, as noted for equation (\ref{eq:shatt_output}) the pad token was a special one mapped to zero and directly integrated in the decoder's architecture, but not as part of the vocabulary.

\paragraph{Image preprocessing.} The training split contained $82,783$ images of which $60,396$ were in horizontal and $22,387$ in vertical orientation. Moreover, these images had $2,159$ different shapes. The most occurring horizontal shapes were $640 \times 480$ and $640 \times 427$ with respectively a count of 17,797 and 10,269 images, whereas for vertical orientation $480 \times 640$ and $427 \times 640$ with 5,754 and 2,989 images. To become invariant on the orientation, I directly resized the images to a quadratic shape \textit{not} keeping the aspect ratio. In contrast to that \citet{xu_show_2015} resized the smallest edge to 256 pixels while keeping the aspect ratio. Then they cropped the center image to receive an input of shape $224\times224\times3$ by following \citet{simonyan_very_2014}. The problem with this procedure was that objects were possibly cut or cropped out of the image. Therefore I deviated from the procedure to keep possible objects fully within the input images. Given this, I made in my experiments fully use of the annotated object bounding boxes.

To match the input shape of the image feature extractor, the images were resized to $448 \times 448$ pixels using nearest neighbor interpolation. This lead possibly to larger distortions for images with edge lengths smaller than 448. However, there are $71,974$ images in the training split which have $448$ or more pixels in width and height representing a total of 86,95\% images. 

\clearpage

The further image preprocessing for the VGG model in Keras was performed in analogy to \citet{simonyan_very_2014}, because the model is a direct conversion of the published weights. They subtracted the mean ImageNet pixel values $p_{m} = (103.939, 116.779, 123.68)$. Thus, I computed $p_{xy} - p_{m}$ and fed the preprocessed images to the VGG-19 model. Finally, I extracted the feature maps as described in section \ref{sec:reimplementation} for the training and validation split.


\begin{table}
	\begin{tabular}{| r | r | r | r | r | r | r | r |}
		\hline
		Lambda    & Epoch & m-loss & m-acc & BLEU-1 & BLEU-2 & BLEU-3 & BLEU-4 \\
		\hline
		0.010 	  &    12 &   3.51 &  0.53 &   69.3 &   51.4 &   36.7 &   26.0 \\
		0.010 	  &    16 &   3.53 &  0.55 &   68.6 &   50.8 &   35.6 &   26.1 \\
		\hline
		0.005 	  &    12 &   2.41 &  0.54 &   \textbf{70.0} &   \textbf{51.9} &   37.1 &   26.2 \\
		0.005 	  &    10 &   2.45 &  0.53 &   69.8	&	51.8 &   \textbf{37.2} &  \textbf{26.6} \\
		\hline
		0.001 	  &    12 &   1.64 &  0.54 &   69.7 &   51.5 &   36.8 &   26.3 \\
		0.001 	  &    11 &   1.66 &  0.54 &   69.4 &   51.6 &   37.0 &   26.4 \\
		\hline
	\end{tabular}
	\caption[The training scores for my ``Show, Attend and Tell'' generator]{Training epochs with the best BLEU scores per $\lambda$ as well as the masked training loss and accuracy. Masking means that predicted words for time-steps after the last word in the ground-truth caption are ignored. The dropout rate is 50\%. }
	\label{table:training_parameters}
\end{table}

\subsection{Training, Hyperparameters and Scores}
\label{sec:training}

As in the work of \citet{xu_show_2015}, I used the Adam optimizer \parencite{kingma_adam:_2014} to minimize the penalized loss function

\begin{equation}
L = -log(p(y|a)) + \lambda \sum_{i}^{L}{(1 - \sum_{t}^{C}{ \alpha_{it} })^2}
\end{equation}

where $p(y|a)$ is the decoder's output as the probability of a caption given a set of extracted image features, $L$ is the number of image features, $C$ is the caption length and $\alpha_{it}$ is the spatial attention for an image region at a time-step or for a specific word prediction. The alpha-regularizer $\lambda$ constraints the caption generator to distribute the spatial attention more equal among the image areas during the whole generation process. \citet{xu_show_2015} noted that this regularizer is important for the resulting overall BLEU score \parencite{papineni_bleu:_2002}, but they mentioned not the exact value to be chosen.

\clearpage

\begin{figure}
	\centering
	\includegraphics[width=\textwidth]{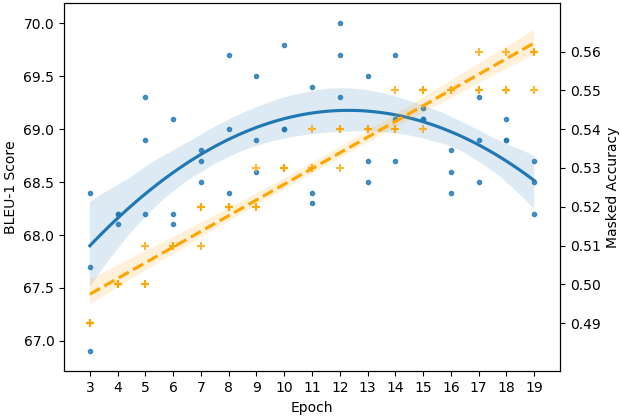}
	\caption[The discontinuity for BLEU scores and accuracy]{The BLEU-1 scores in blue exemplary chosen for the three $\lambda$-models and plotted with the masked accuracy in percent. The scores are fitted with a second order polynomial while the accuracy fit in orange is linear. The increase in score shows to disconnect with accuracy increment in later epochs. The regularizer effects are exemplary shown in Appendix \ref{AppendixA}.}
	\label{fig:plot_bleu_vs_accuracy}
\end{figure}

Therefore, I examined various training runs with batch size 64 and different alphas as shown in Table \ref{table:training_parameters}. The dropout rate was fixed to 50\%. After each epoch I calculated the BLEU-1,2,3,4 scores regarding the validation split from \citet{karpathy_deep_2017} as referred to in the work of \citet{xu_show_2015}. For this I instantiated a sibling model with an arg-max sampler on the decoder's output vector. This means, that I chose the highest probably word as the output word at a specific time-step for validation and stopped the caption production on the end symbol.

\citet{xu_show_2015} used early stopping as an additional regularization technique, because they found that accuracy and BLEU score discontinue to correlate in later epochs. As exemplary shown for BLEU-1 in Figure \ref{fig:plot_bleu_vs_accuracy}, my results revealed the same behavior with a decline in scores after around 12 epochs. 

\clearpage

In difference to \citet{xu_show_2015} who probably chose the last model, I picked the model with the best BLEU-4 score over the epochs. As we can see, I was partially able to increase on the reported scores as my best model achieved $69.8$, $51.8$, $37.2$, $26.6$ in BLEU-1,2,3,4 respectively. \citet{xu_show_2015} reported $70.7$, $49.2$, $34.4$, $24.3$ which is a difference of $-0.9$, $+2.6$, $+2.8$, $+2.3$ in score. 

\section{Spatial Attention Interface Methods}
\label{sec:control_mechanisms}

Now, given the working assumptions from chapter \ref{Chapter2}, I ask, whether through the spatial attention of a caption generator its output is controllable and if this connection could be reversed. It this holds true, then a trained image captioning model should be able to describe individual objects in a complex scenery. For example, when forcing the spatial attention into a specific region of an image. 

These reasonable reactions to manipulation in the attention could then be used to condition a captioning model and to generate captions that are guided by external spatial attention that had been produced from another model in the scope of other tasks like visual question answering. The benefit of using spatial attention as an interface is that even completely different architectures can rely on the simple constraints of $\alpha \in \mathbb{R}^{L}$, $\alpha \in (0,1)$ and $\sum{\alpha} = 1$ which are also to some extent human interpretable. 

In this section I exemplary examine the generated captions under a fixed spatial attention and then introduce three different interface methods to conduct this attention manipulation. 

\subsection{Testing the Effect of Spatial Attention Fixation}

First, I tested manually, whether the trained captioning model from before was actually reacting to changes in its spatial attention. Therefore, I chose an image from the validation split of the same dataset that contained multiple objects: a sailing boat in the background, an ocean wave in the middle and a surfer in the foreground. 

\clearpage

\begin{figure}
	\centering
	\includegraphics[width=.6\textwidth]{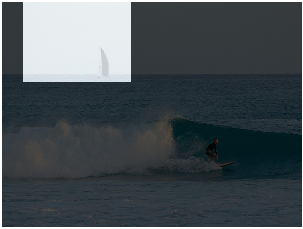}
	\caption[Testing the effect of spatial attention fixation]{The spatial attention is manipulated to be of an uniform value in the highlighted region while the attention is set to zero everywhere else. The caption generator produces: \\ \textit{a man is standing in the water with his surfboard}}
	\label{fig:attention_shift_surfer}
\end{figure}

For this image, I defined a spatial attention matrix of size $14 \times 14$, which corresponds to the size of the extracted image feature vector. The spatial attention matrix was set everywhere to zero except for a $5 \times 5$ large window. Within this attention region an equal value of $\frac{1}{25}$ was placed. Given such an attention matrix, the constraints $\alpha \in (0,1)$ and $\sum_{i}^{L}{\alpha_{i}} = 1$ were satisfied. This window of attention was then shifted nine times over the top, middle and bottom of the image. The second top shift is exemplary shown in Figure \ref{fig:attention_shift_surfer}.

Although, the produced captions are rather monotonic in nature, they show a tendency based on the position of the spatial attention window. The upper captions are referring to \textit{a man} \textit{standing} with his or in front of a \textit{surfboard}. The middle and bottom captions are more related to the actual action taking place by describing \textit{a surfer is riding a wave}. This is plausible, because surfer and wave are within the window of attention. Similar observations were made on further randomly sampled images, thus I conclude that the shifting spatial attention is indeed having an effect on the produced caption. In the following, I describe three controlling mechanisms based on this idea, which I will apply in the experiments to evaluate whether the made observations are indicating a systematic behavior of the underlying caption generator.

\clearpage
\subsection{Unlimited Step-wise Fixed Attention}

\begin{figure}
	\centering
	\includegraphics[width=\textwidth]{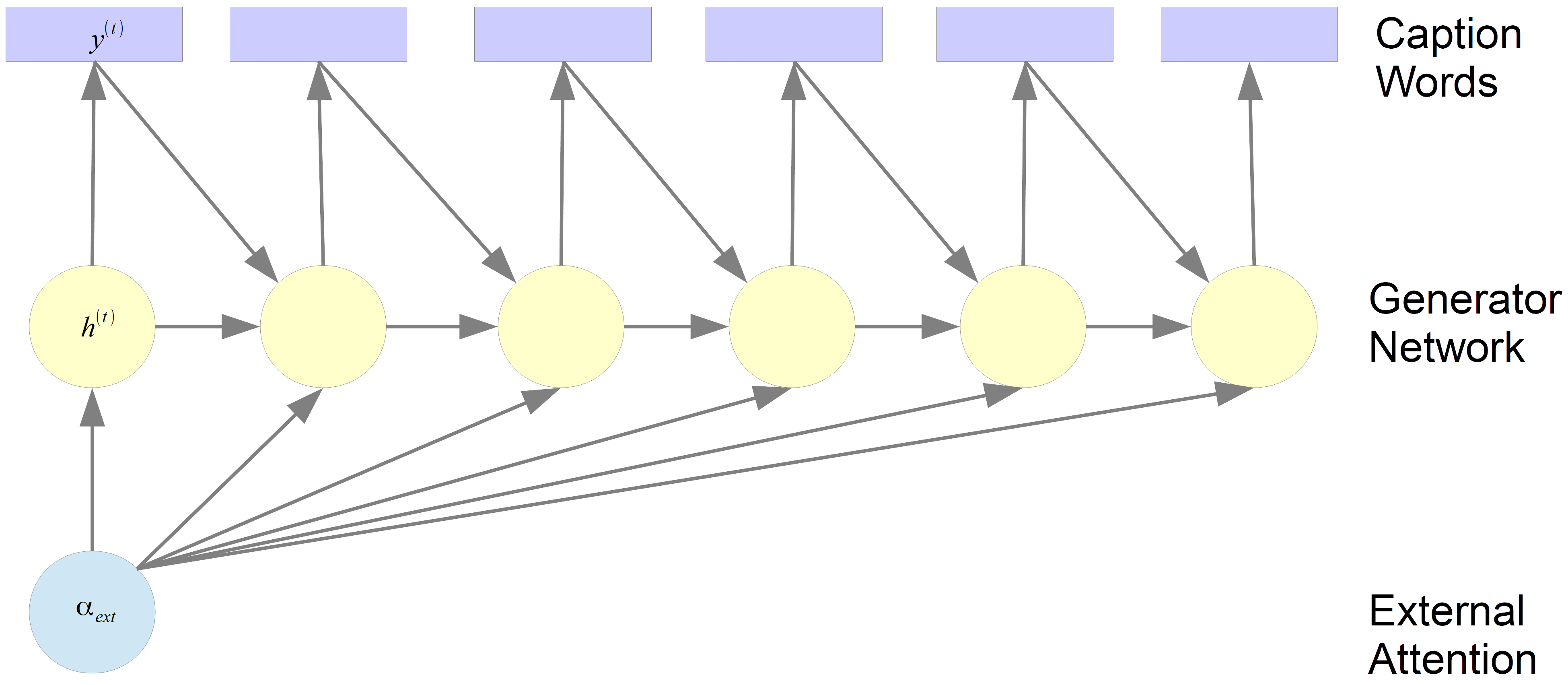}
	\caption[An architectural view on the unlimited step-wise fixed attention]{An architectural view on the unlimited step-wise fixed attention. The externally provided spatial attention vector in blue is fed to the hidden state computation in yellow for all time-steps during the generation of the caption words shown in purple. The attention network of the generator is disabled.}
	\label{fig:methods_fixed}
\end{figure}

For the unlimited step-wise fixed attention mechanism, I fed an external spatial attention vector at each time step to the trained image captioning model for the whole caption generation process as shown in Figure \ref{fig:methods_fixed} while the model's own predicted attention is dismissed. The external spatial attention vector is kept constant, thus at each time-step the caption generator's context vector was computed following

\begin{equation}
z^{(t)} = \alpha_{\text{ext}} \text{A}
\end{equation}

so that $z^{(t)} \in \mathbb{R}^{D} $, because the extracted image features are $A \in \mathbb{R}^{L \times D}$  with $D$ as the number of image feature maps and $L$ as the amount of spatial image features per feature map. Thus, the external attention vector is $\alpha_{\text{ext}} \in \mathbb{R}^{L}$.

\clearpage
\subsection{Limited Step-wise Fixed Attention} 

\begin{figure}
	\centering
	\includegraphics[width=\textwidth]{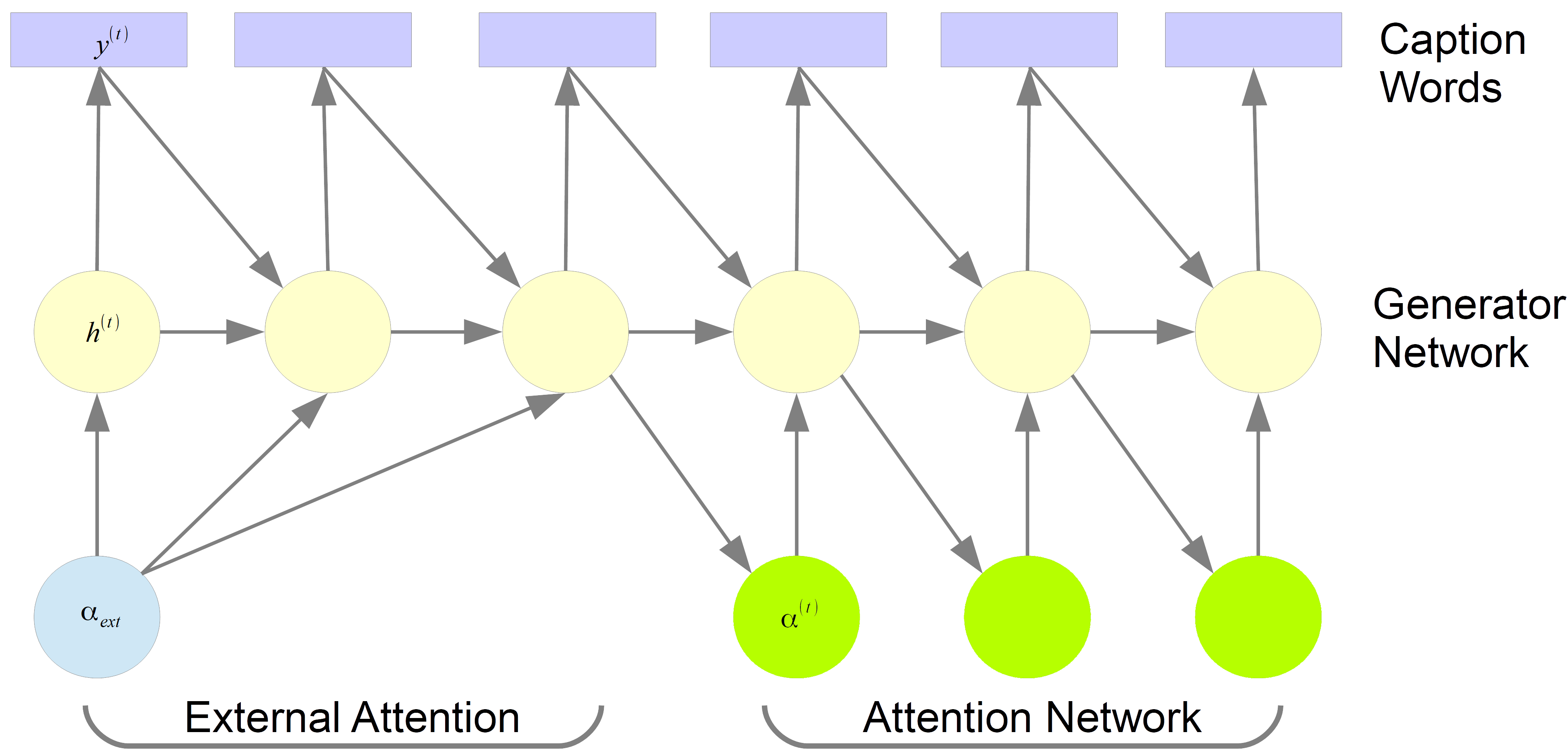}
	\caption[An architectural view on the limited step-wise fixed attention]{An architectural view on the limited step-wise fixed attention. The externally provided spatial attention vector in blue is fed to the hidden state computation in yellow for the first \textit{i} time-steps during the generation of the caption words shown in purple. Meanwhile the attention network is disabled. After the \textit{i}-th step, the attention network is fully in charge again.}
	\label{fig:methods_semi}
\end{figure}

For the limited step-wise fixed attention, I fed an external spatial attention vector for the first $i$ time steps as shown in Figure \ref{fig:methods_semi}. After the $t_{i}$ time step, the model is again ``free to choose'' the spatial attention depending on its state and the previous word using the attention network. Therefore, the caption generator's context vector was computed following

\begin{equation}
	z^{(t)} = 
	\begin{cases}
	\alpha_{\text{ext}} \text{A} \textnormal{ if } t \leq i\\
	\alpha^{(t)} \text{A}  \textnormal{ otherwise }
	\end{cases}
\end{equation}

so that $z^{(t)} \in \mathbb{R}^{D} $, because the extracted image features are $A \in \mathbb{R}^{L \times D}$ with $D$ as the number of image feature maps and $L$ as the amount of spatial image features per feature map. Thus, the external attention vector is $\alpha_{\text{ext}} \in \mathbb{R}^{L}$ and the model's predicted attention vector is $\alpha^{(t)} \in \mathbb{R}^{L}$ as well.

\clearpage
\subsection{Step-wise Additive Attention} 

\begin{figure}
	\centering
	\includegraphics[width=\textwidth]{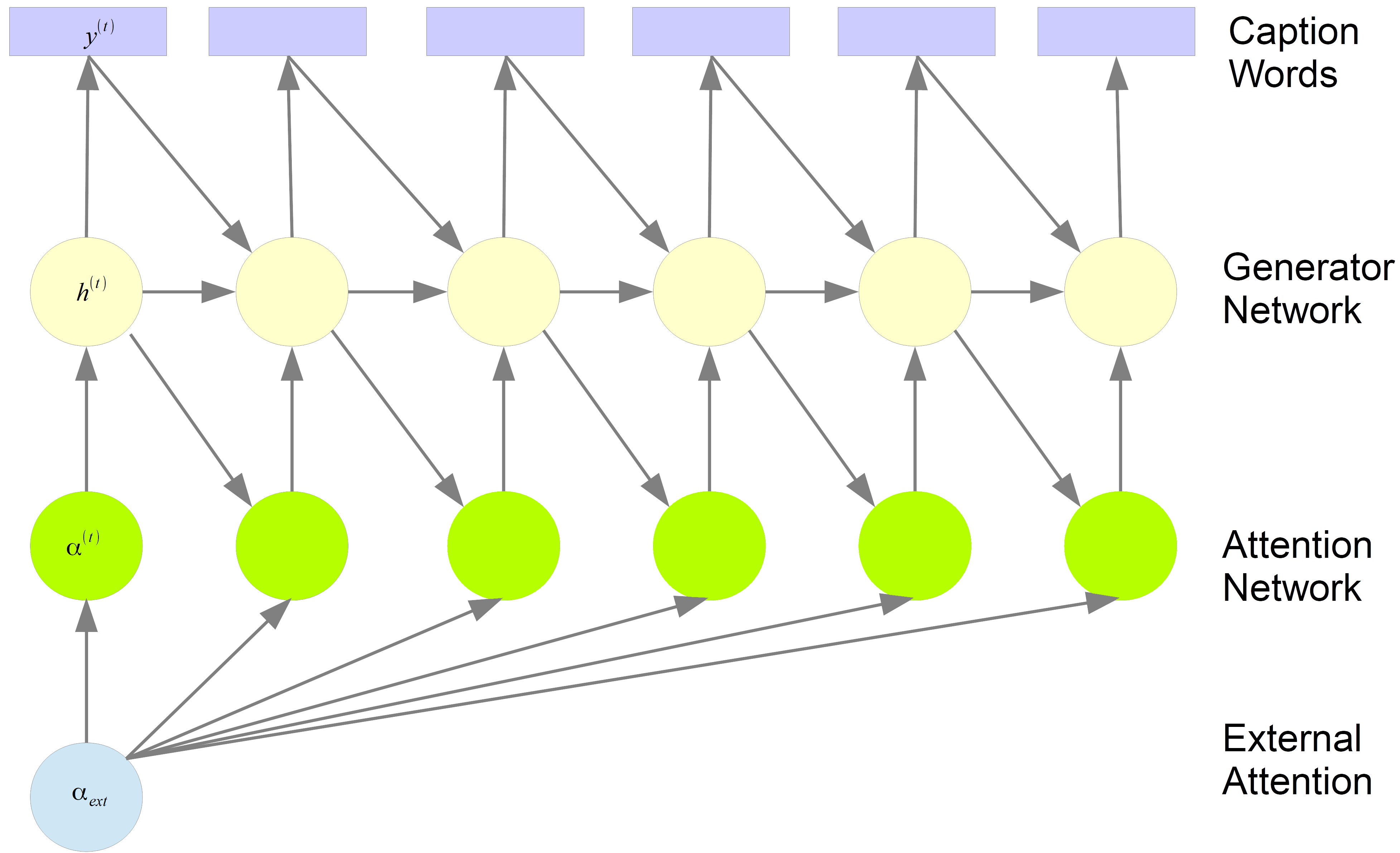}
	\caption[An architectural view on the step-wise additive attention]{An architectural view on the  step-wise additive attention. The externally provided spatial attention vector in blue is added to the attention network's predicted one and fed to the hidden state computation in yellow for all time-steps during the generation of the caption words shown in purple. }
	\label{fig:methods_additive}
\end{figure}

For the step-wise additive attention, I fed an external spatial attention vector at each time-step to the model as shown in Figure \ref{fig:methods_additive}, but in contrast to the fixed attention, I compute the weighted sum of the model's predicted and the external attention. The weighting scalar $\phi$ defines the influence of the external attention and thus the generator's context vector is computed with 

\begin{equation}
	z^{(t)} = (\frac{\alpha^{(t)} + \phi \alpha_{\text{ext}}}{\phi + 1}) \text{A}
\end{equation}

so that $z^{(t)} \in \mathbb{R}^{D} $, because the extracted image features are $A \in \mathbb{R}^{L \times D}$ with $D$ as the number of image feature maps and $L$ as the amount of spatial image features per feature map. Thus, the external attention vector is $\alpha_{\text{ext}} \in \mathbb{R}^{L}$ and the model's predicted attention vector is $\alpha^{(t)} \in \mathbb{R}^{L}$ as well.


\chapter{Experiment 1: Interface the Generator Using Bounding Boxes} 

\label{Chapter4} 

Now, given the previous remarks, I assumed that a sufficiently well trained captioning system is capable of talking about a variety of objects and object configurations. Thus, in this experiment I examined, if the output of a standard caption generator is predictable and in such a way controllable that I could interfere with its spatial attention to produce a caption that contains object annotations which are known for an image. Therefore, I manipulated the trained caption generator from section \ref{sec:training} to evaluate the high level relationship between the model's attention and the produced captions. 

\section{Experimental Setup}
\label{sec:experiment1_setup}

As an exemplary collection of object annotations for images I used a standard image detection dataset. Given this, I derived spatial attention vectors from the object annotation bounding boxes. Then the image captioning model had to produce what I call a \textit{box caption} given such a spatial attention vector. 

The generated box captions were supposed to contain words which refer to the annotated object categories within the images. To provide insights with respect to the effectiveness and sensitivity of the model, I introduced two measurement methods regarding the spatial attention manipulation techniques described in section \ref{sec:control_mechanisms}.

\clearpage

\subsection{An Object Detection Dataset} 

For this experiment, I used the MSCOCO validation split of the \emph{Detection Challenge 2015}. This was in particular useful, because the caption generator had been trained on the same domain of images, but without the validation images. However, the model had to produce captions about the validation images to compute the BLEU scores during the training. Here, I could use these captions as a reference for ``what would have been normally produced'' and thus compare them with the box captions, which were generated in this experiment, to study the relative effect of the attention manipulations. 

The MSCOCO dataset \parencite{lin_microsoft_2014} provided a varying amount of bounding boxes for each of the 40,504 validation images. The boxes framed distinct, but possibly overlapping objects that were each annotated with one of 80 object categories. The most objects were given for the categories \textit{person}, \textit{shoe}, \textit{window}, \textit{car} and \textit{hat}. Given these, there was an total amount of 291,875 bounding boxes in the validation split. The bounding boxes were defined as rectangles given a width $b_{w}$, height $b_{h}$ and the xy-coordinates $b_{xy}$ of the left upper corner in the according image. The mean/median width and height was 104.46/54.45 and 108.04/62.85 respectively. 

I discarded all bounding boxes that are smaller than the median size, because I assumed that the model less likely attends to small objects in the images and because I used nearest neighbor down-sampling which keeps the sharp box edges during projection into the attention space. Therefore, the smallest possible bounding box had to be at least of size $32 \times 32$, so that it could be represented within the $14 \times 14$ spatial attention vector as a single entry. As a result, there were 2,865 images that were ignored in this experiment. None of their bounding boxes were larger than the median. In the end, there were 117,798 remaining bounding boxes for 37,639 images from which I derived the spatial attention vectors.

\clearpage

\subsection{Using Bounding Boxes as External Spatial Attention}

I constructed for each bounding box a 196-dimensional spatial attention vector. To achieve this, I first initialized for each bounding box a zero matrix with the width and height of the image in which the bounding box was placed. This step was necessary, because the image sizes were varying as described in section \ref{sec:mscoco}. The variation introduced an image dependent scaling factor of the bounding box rectangles and as such determined which entries of the spatial attention vector were involved. This zero matrix corresponded to a black one-channel image on which I ``drew'' the bounding box as a white rectangle with the boxes upper-left corner at the pixel $p_{xy} = b_{xy}$ ranging from $p_{x}$ to $p_{x} + b_{w}$ and from $p_{y}$ to $p_{y} + b_{h}$. The resulting image matrix had $p_{xy} = 255$ everywhere within the bounding box and otherwise $p_{xy} = 0$. Then I resized the bounding box image to the shape $14 \times 14$ using nearest neighbor down-sampling not keeping the aspect ratio. Finally, the matrix was flattened to a 196-dimensional vector. In contrast to the manually created attention vectors used in section \ref{sec:control_mechanisms}, the down-sampling procedure could have introduced artifacts in the resulting spatial attention vectors.

During the experimental runs, I loaded for each image in the validation split all according spatial attention vectors as a single batch. Then I interpolated for each vector the pixels values following $[0,255] \rightarrow [0,1]$ to guarantee that $\alpha \in [0,1]$. In addition, I applied the softmax function on the flatten vector, so that also $\sum{\alpha}_{i} = 1$ is guaranteed like in the implementation of \citet{xu_show_2015}. An important detail is that the softmax result is small, but nowhere zero. Thus, the model is still allowed to include image aspects outside the boxes for the caption generation. Finally, I applied the bounding boxes as external spatial attention using the interface methods in the following configurations:

\begin{itemize}
\item[I.] \textbf{Unlimited step-wise fixed attention.} I fed the spatial attention vector at each time step to the model for the whole caption generation process and dismissed the model's one. The expectation was that the model would be highly forced to tell something about the bounding box objects, when the model has seen enough similar samples during the training. See Figure \ref{fig:dog_fixed_sum} for a visual example.

\item[II.] \textbf{Limited step-wise fixed attention.} I fed the spatial attention vector for the first $i = \{3,6,9\}$ time steps which were empirically chosen. See Figure \ref{fig:dog_semi_fixed_sum_multiple_iterations_on_bicycle} for a visual example. I used this setup, because in captions the object description usually come first e.g. \textit{a dog sitting on a couch}, if the bounding box includes a dog. 

\item[III.] \textbf{Step-wise additive attention.} At each time step, the spatial attention vector was added to the one predicted by the model. I used the weighting factor values $\phi = \{1,2,3\}$ to control the weight of the externally induced attention. See Figure \ref{fig:dog_dynamic_sum_multiple_weights_on_bicycle} for a visual example. This techniques still allowed the model to use the own predicted attention over the whole generation process. However, when larger weights are put on the external attention, then this should lead to similar results like with the unlimited fixed attention method. Besides, when we interpret the external attention as a guidance for the generator, then the additive method is the most interesting one for interaction with another model. 
\end{itemize}

\subsection{Statistics}

\paragraph{Result Captions.} The trained image captioning model had to produce what I call a \textit{box caption} for each constructed attention vector with each of the attention control mechanisms. The box caption as the output of the model under a specific forcing method indicated the impact of the spatial attention manipulation in relation to the normally produced caption. 

Therefore, I included the captions generated during training for computation of the BLEU scores on the validation set as the normally generated caption and call them \textit{self-attending caption} in the following, because the attention is ``freely chosen'' by the model during the whole caption generation process.

In addition, I indicated whether the model's changes in caption generation were related to specific attention forcing methods or a method unrelated phenomenon. To do so, I let the model produce a \textit{control caption} where the spatial attention had been distributed uniformly over the whole image with a value of $\frac{1}{196}$. This made it possible to study the effect of the individual forcing methods in relation to a naive manipulation of the spatial attention. 

\clearpage

\begin{figure}
	\centering
	\includegraphics[width=0.6\textwidth]{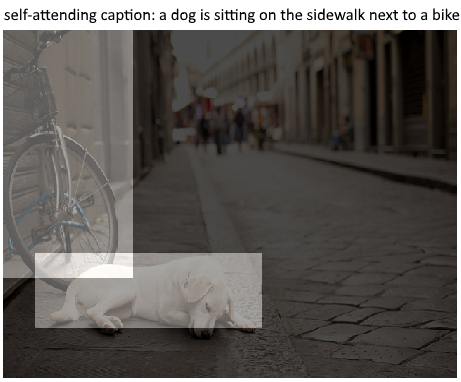}
	\caption[Attention maps focusing on a dog and a bicycle]{An image from the validation set with attention maps focusing on a dog and a bicycle. Above the caption that had been produced during validation of the caption generator.}
	\label{fig:validation_image_with_boxes}
\end{figure}

\paragraph{Degree of  Sensitivity.} The model was supposed to react to changes in its spatial attention. I suggested the \textit{degree of sensitivity} as a measurement for this capability. Here, I quantified how often the resulting box captions deviated from the normally generated self-attending caption for an image. I counted the model sensitive to a spatial attention vector, when the resulting box caption differed in at least a single word e.g. \textit{a dog is \underline{laying} on the sidewalk next to a bike}, although this caption did not tell something new about the image. Furthermore, I counted the model sensitive to a specific forcing method, when the box caption differed from the control caption in at least a single word. 

For example, there might be two bounding boxes and thus two spatial attention vectors fed to the model for an image like in Figure \ref{fig:validation_image_with_boxes}. Now, a spatial attention focus might have resulted into a box caption that is the same as the self-attending caption, even when put on the bicycle, because the model was biased towards the dog given such an image (as I stated above, the spatial attention is nowhere zero). On the other hand, when the spatial attention was put on the dog given its bounding box, then the model might have produced the same caption as during the validation run. Such a result counted as \textit{not-sensitive} into the sensitivity of the model although it was a correct statement. 

\clearpage

\paragraph{Degree of Controllability.} The main objective for applying the external spatial attention using bounding boxes was to let the model produce box captions. These box captions were supposed to refer to objects within the bounding boxes from which the spatial attention vectors were derived. Thus I suggested the \textit{degree of controllability} as a measurement by checking that the box caption included the according box category. This metric provided a lower boundary on the performance of the attention control methods, since a freely generated caption is compared with a restricted list of classes.

For example, one of these eighty box categories was \textit{bicycle}. Now, when the spatial attention focus was set on the bicycle in Figure \ref{fig:validation_image_with_boxes}, then the model might have produced a box caption that included the category word. If this happened, I counted the model as controllable with respect to the bounding box spatial attention vector. Nevertheless, such an exact word matching was very restrictive, since we would intuitively also relate the term \textit{bike} as correct. 

Therefore, I evaluated the resulting box captions also with respect to the k-nearest word neighbors of a category word. I called this \textit{k@5} for including the five nearest-neighbors in the evaluation. These word neighbors were retrieved from the model by determining the learned word embedding vectors for the whole vocabulary. Then I took for each category word the k-nearest words with respect to the cosine distance. For compound word categories like \textit{fire hydrant}, I determined the k-nearest neighbors individually and then mapped them to the same original category. This was necessary, because the vocabulary was composed only of individual words. 

For the trained model, the five nearest neighbors for bicycle also included \textit{bike}, \textit{motorcycle}, \textit{bicycles} and \textit{bikes}. Given these, the model was also counted as controllable under k@5, when the box caption contained bike. However, the self-attending caption already included bike as shown in Figure \ref{fig:validation_image_with_boxes}, so that I was not able to conclude, whether the forcing method had an impact on the caption generation. Therefore, I additionally introduced the \textit{distinct subset} for which I discarded box captions from the evaluation in cases, where the model already mentioned box categories (k@1) or its neighbors (k@5) in the self-attending caption.

\clearpage

\section{Results}

\paragraph{Qualitative Results.} Here, I give exemplary results for the image shown in Figure \ref{fig:validation_image_with_boxes} to provide a better understanding of the quantitative evaluation. The caption generator was producing under the unlimited step-wise attention method the following box and control captions

\begin{itemize}
\item 	\textbf{box caption 1 (dog)\hspace{1.1cm}:} a \textit{dog} is laying on a leash on a leash
\item 	\textbf{box caption 2 (bicycle)\hspace{.5cm}:} a \textit{bicycle} is parked next to a \textit{bike} rack
\item 	\textbf{control caption (uniform)\hspace{.0cm}:} a large white \textit{dog} is standing in a sidewalk
\item 	\textbf{self-attending (normal)\hspace{.4cm}:} a \textit{dog} is sitting on the sidewalk next to a \textit{bike}
\end{itemize}

Given these results, the degree of sensitivity was one hundred percent for both general and method sensitivity, because the box captions deviated from the self-attending and from the control caption. In terms of controllability both captions counted towards the general subset, because box caption 1 contained the box category \textit{dog} and box caption 2 included the category word \textit{bicycle} (therefore, this is also the case for k@5). 

For the distinct subset only box caption 2 was counted and the box caption 1 was discarded, because the self-attending caption already mentioned a dog. Furthermore, the box caption 2 was additionally discarded, when incorporating the nearest neighbors, because they included \textit{bike} and this was also already mentioned in the self-attending caption.

As a negative example, the model was producing under the step-wise additive attention with a weighting factor of 1 the same sentence for box 1, box 2 and the control caption: \textit{a dog is sitting on a bench}. Therefore, these captions would have counted into the general sensitivity, but not as method sensitive, because the box captions were the same as the control caption. 

In the negative example, only box caption 1 was counting into the controllability measure by containing \textit{dog} as the correct box category. The box caption 2 was not mentioning \textit{bicycle}, which would have been the correct category. As a result, also in the distinct subset both box captions were not included.



\clearpage

\paragraph{Quantitative Results: Degree of Sensitivity.}

\begin{table}
	\centering
	\begin{tabular}{| l | c | c |}
		\hline
		&  \multicolumn{2}{c|}{Sensitivity} \\
		& general (diff) & method (diff) \\
		\hline
		unlimited      & 88.68 (0.55) & \textbf{52.65} (0.54)  \\
		\hline  
		limited-3      & 85.23 (0.55) & 35.20 (0.55) \\
		limited-6      & 87.90 (0.56) & 46.49 (0.55) \\
		limited-9      & \textbf{88.88} (0.55) & 51.81 (0.54) \\
		\hline  
		additive-1     & 85.51 (0.54) & 33.43 (0.51) \\
		additive-2     & 87.26 (0.54) & 41.25 (0.53) \\
		additive-3     & 85.49 (0.54) & 44.29 (0.52) \\
		\hline             
	\end{tabular}
	\caption[Results for the degree of sensitivity]{The degree of sensitivity as the percentage of 117,798 box captions which deviate from the control or self-attending caption in at least a single word indicating method or general sensitivity respectively. The differentness for each evaluation set is given in parentheses as  mean word error rate.}
	\label{table:sensitive}
\end{table}

As shown in Table~\ref{table:sensitive}, the caption generator had the highest general sensitivity for the limited step-wise fixed attention method with nine iterations where 88.88\% of the box captions differed from the self-attending caption in at least a single word. The smallest general sensitivity was given during application of the limited step-wise fixed attention method with three iterations resulting into 85.23\%. 

The highest degree of method specific sensitivity was measured for the unlimited fixed attention as depicted in Table~\ref{table:sensitive}. Using this method, 52.65\% of the box captions differed to the control caption in at least a single word. The smallest method specific sensitivity was measured during the step-wise additive attention with a weighting factor of 1 resulting in 33.43\%. 

In addition, I computed the word error rate (WER) to indicate how much the captions differed from the self-attending or control caption. The mean WER scores ranged from 0.51 to 0.56 for general and method sensitivity. The highest rates of 0.56/0.55 were given for the limited step-wise fixed attention with six iterations while the lowest rates were observable given the additive attention configurations with 0.54/0.51. We can say that on average the box captions differed in approximately every second word in comparison to the self-attending or control captions.

\paragraph{Quantitative Results: Overall Controllability.}

\begin{table}
	\centering
	\begin{tabular}{| l | r | r | r | r |}
		\hline
		& \multicolumn{2}{c}{Controllable} & \multicolumn{2}{c|}{and distinct}  \\
		&  k@1  & k@5   &  k@1  & k@5 \\
		\hline
		unlimited      & \textbf{28.56} & \textbf{58.17} &\textbf{ 9.00} & \textbf{21.39} \\
		\hline  
		limited-3      & 26.36 & 50.84 & 6.89 & 15.24  \\
		limited-6      & 27.69 & 52.75 & 8.21 & 17.86 \\
		limited-9      & 27.32 & 52.94 & 7.85 & 18.03  \\
		\hline  
		additive-1     & 25.86 & 52.89 & 6.27 & 17.23  \\
		additive-2     & 26.98 & 52.28 & 7.26 & 16.70  \\
		additive-3     & 27.35 & 53.83 & 7.33 & 18.69  \\
		\hline             
	\end{tabular}
	\caption[The results for the overall degree of controllability]{The degree of controllability as the percentage of box captions containing their category in relation to all 117,798 box captions. The degree for the distinct share is based on 87,033 (k@1) or 58,407 (k@5) box captions where the self-attending caption is \textit{not} already including the box category.}
	\label{table:controlable}
\end{table}

Table~\ref{table:controlable} shows the highest degree of controllability for the unlimited step-wise attention fixation which resulted in 28.56\% (k@1) of the cases in a box caption that included the box category. For compound categories both words had to be included. The least controllability with exact word matching was shown by the additive step-wise attention using a weighting factor of 1 with 25.86\% (k@1).

When the box caption was allowed to include the five nearest category words, then the highest degree of controllability was given for the unlimited step-wise attention showing 58.17\% (k@5). For compound categories, the two nearest words were chosen and the compound category word itself. The least controllability was here given by the limited step-wise attention fixation for three iterations with 15.24\% (k@5).

Table~\ref{table:controlable} shows for the distinct subset that the highest controllability was given for the unlimited step-wise fixed attention, where in 9.0\% (k@1) of the cases the resulting caption included the box category. In 21.39\% (k@5) of the cases the caption included at least one of the nearest words, when the spatial attention was focusing on something new or not mentioned before in the image. The least controllability for the distinct subset was given by the additive attention with weight 1 and limited step-wise attention fixation with three iterations respectively having 6.27\% (k@1) and 15.24\% (k@5).

\clearpage

\begin{table}
	\centering
	\begin{tabular}{| r || l | r | r || l | r | r |}
		\hline
		& \multicolumn{3}{c||}{Matches with k@1} & \multicolumn{3}{c|}{Matches with k@5} \\
		\hline
		Pos. & Category    & Rel. & Abs.   & Category        & Rel.  & Abs. \\
		\hline
		1 & cat            & 89 \% & 1,370 &  zebra         & 98 \% &  1,386 \\
		2 & train          & 85 \% & 1,209 &  cat           & 94 \% &  1,445 \\
		3 & pizza          & 85 \% & 1,205 &  $\uparrow_{12}$ giraffe       & 94 \% &  1,409 \\
		4 & toilet         & 85 \% & 1,067 &  train         & 89 \% &  1,264 \\
		5 & clock          & 85 \% &   628 &  $\uparrow_{41}$ elephant      & 87 \% &  1,169 \\
		6 & fire hydrant   & 83 \% &   332 &  pizza         & 86 \% &  1,221 \\
		7 & zebra          & 77 \% & 1,094 &  bear          & 86 \% &    362 \\
		8 & bear           & 77 \% &   323 &  toilet        & 85 \% &  1,067 \\
		9 & $\uparrow_{2}$ dog            & 76 \% & 1,175 &  clock         & 85 \% &    630 \\
		10 & $\uparrow_{4}$ sheep         & 76 \% &   950 & fire hydrant   & 85 \% &    339 \\
		\hline
		\hline
		11 & ... & ... & ... & $\downarrow_{2}$ dog         & 84 \%  & 1,289  \\
		14 & ... & ... & ... & $\downarrow_{4}$ sheep       & 78 \%  & 977  \\
		15 & $\downarrow_{12}$ giraffe & 69 \%  & 1,030 & ... & ... & ... \\
		36 & ... & ... & ... & $\uparrow_{38}$ person  & 60 \% & 22,338 \\
		46 & $\downarrow_{41}$ elephant    	 & 36 \%  &  485 & ... & ... & ...  \\
		67 & $\uparrow_{6}$ baseball glove   & 10 \%  &   12 & ... & ... & ...  \\
		68 & $\uparrow_{4}$ orange       	 &  9 \%  &   80 & $\uparrow_{7}$ potted plant    & 19 \% &  273 \\
		\hline
		\hline
		71 & apple         & 7 \% &     41 & apple          &  14 \% &     88 \\
		72 & backpack      & 6 \% &     55 & $\downarrow_{4}$ orange         &  11 \% &     96 \\
		73 & toaster       & 4 \% &      1 & $\downarrow_{6}$ baseball glove &  11 \% &     14 \\
		74 & $\downarrow_{38}$ person        & 2 \% &    877  & backpack       &   8 \% &    71 \\
		75 & $\downarrow_{7}$ potted plant  & 2 \% &     22  & dining table   &   5 \% &   190 \\
		76 & carrot        & 2 \% &     19 & toaster        &   4 \% &      1 \\
		77 & dining table  & 0 \% &     12  & carrot         &   3 \% &    22 \\
		78 & handbag       & 0 \% &      0  & sports ball    &   3 \% &     3 \\
		79 & sports ball   & 0 \% &      0 & handbag        &   2 \% &     20  \\
		80 & hair drier    & 0 \% &      0 & hair drier     &   0 \% &      0 \\
		\hline             
	\end{tabular}
	\caption[The degree of controllability by category for the whole split]{The degree of controllability by box category for the \textit{whole split}. The top and bottom ten box categories are listed regarding the percentage of matches. A match means that a box captions includes the category word (k@1) or one of the nearest neighbors (k@5). The arrows indicate the amount of position in an upwards or downwards direction, when switching between k@1 and k@5. There are in total 117,798 box captions and 80 box categories. For category-wise total amounts see Appendix \ref{AppendixC}.}
	\label{table:controlability_per_category}
\end{table}

\clearpage

\paragraph{Quantitative Results: Category-wise Controllability.} For the discussion, I provide  category-wise results exemplary for the unlimited step-wise fixed attention, because that method was working best based on the overall results.

Table \ref{table:controlability_per_category} shows that the caption generator produced in 89\% of the cases a caption with the category word \textit{cat}, when one of the 1,542 spatial attention vectors referring a cat was provided. This rate increases to 94\% when also word neighbors were allowed, but then the best box category was \textit{zebra} with 98\%. The categories cat, train, pizza, toilet, clock, fire hydrant, zebra and bear occurred in the top ten for both exact and neighbor matching. For both matching methods five animal box categories were present in the top ten.
The worst results showed the generator on exact matching for the categories dining table, handbag, sports ball and hair drier with 0\% matches, followed by person, potted plant and carrot with 2\% each. These numbers slightly increased, when neighbor words were allowed. Still, sports ball, handbag and hair drier stayed last with 3\%, 2\% and 0\%. The categories apple, backpack, toaster, carrot, dining table, handbag, sports ball and hair drier occurred in the bottom ten for both exact and neighbor matching. The categories person and potted plant were in the bottom ten only for exact matching.

The bottom ten looks similar for the category-wise controllability on the distinct subset as shown in Table \ref{table:controlability_per_category_distinct}. The worst results were given for exact matching on the categories dining table, handbag, sports ball and hair drier. These categories resulted in 0\% matches, followed by person and carrot with 1\% each. The numbers slightly increased, when neighbors were allowed. However, hair drier stayed last with no matches, followed by sports ball, handbag and dining table with 1\% each. The categories person and spoon were in the bottom ten only for exact matching. 
The best results on the distinct subset was measured for the category zebra. Here, in 67\% of the cases a caption was produced that contains the category word. The relevant matches increased to 68\%, when also word neighbors were incorporated. The box categories zebra, pizza, motorcycle, parking meter, laptop, toilet and fire hydrant occurred in the top ten for both exact and neighbor matching. In addition, there were three animal categories in the top ten: zebra and giraffe for exact matching as well as zebra and horse for neighbor matching.

\clearpage

\begin{table}
	\centering
	\begin{tabular}{| r || l | r | r || l | r | r |}
		\hline
		& \multicolumn{3}{c||}{Matches with k@1} & \multicolumn{3}{c|}{Matches with k@5} \\
		& \multicolumn{3}{c||}{on distinct subset} & \multicolumn{3}{c|}{on distinct subset} \\
		\hline
		Pos. & Category    & Rel. & Abs.  & Category        & Rel.  & Abs.  \\
		\hline
		1 & zebra           & 67 \% &  275 & zebra           & 68 \% &   28 \\
		2 & pizza           & 48 \% &  113 & $\uparrow_{17}$ tennis racket   & 56 \% &  198 \\
		3 & motorcycle      & 46 \% &  356 & motorcycle      & 54 \% &  374 \\
		4 & $\uparrow_{29}$ giraffe         & 45 \% &  227 & $\uparrow_{14}$ airplane        & 54 \% &  184 \\
		5 & parking meter   & 44 \% &   23 & pizza           & 52 \% &  114 \\
		6 & laptop          & 43 \% &  193 & fire hydrant    & 44 \% &   34 \\
		7 & toilet          & 42 \% &   74 & toilet          & 42 \% &   74 \\
		8 & fire hydrant    & 42 \% &   39 & laptop          & 41 \% &  154 \\
		9 & $\uparrow_{3}$ clock           & 38 \% &   51 & parking meter   & 41 \% &   16 \\
		10 & $\uparrow_{8}$ skateboard     & 36 \% &   69 & $\uparrow_{12}$ horse           & 38 \% &   81 \\
		\hline
		\hline
		12 & ... & ... & ... & $\downarrow_{3}$ clock           & 37 \% &   48 \\
		18 & $\downarrow_{14}$ airplane        & 31 \% &  169  & $\downarrow_{8}$ skateboard      & 36 \% &   57 \\
		19 & $\downarrow_{17}$ tennis racket   & 31 \% &  160  & ... & ... & ... \\
		22 & $\downarrow_{12}$ horse           & 30 \% &  182  & ... & ... & ... \\
		23 & ... & ... & ... & $\uparrow_{52}$ person          & 32 \% & 5409 \\
		33 & ... & ... & ... & $\downarrow_{29}$ giraffe         & 29 \% &   14 \\
		58 & ... & ... & ... & $\uparrow_{15}$ spoon           & 17 \% &   56 \\
		62 & $\uparrow_{9}$ cup             & 9 \% &  129  & ... & ... & ... \\
		70 & $\uparrow_{4}$ orange          & 6 \% &   41  & ... & ... & ... \\
		\hline
		\hline
		71 & backpack        & 6 \% &   46 & $\downarrow_{9}$ cup             & 9 \% &  128 \\
		72 & toaster         & 6 \% &    1 & backpack        & 7 \% &   54 \\
		73 & $\downarrow_{15}$ spoon           & 5 \% &   19 & potted plant    & 6 \% &   62 \\
		74 & potted plant    & 2 \% &   20 & $\downarrow_{4}$orange          & 6 \% &   46 \\
		75 & $\downarrow_{52}$ person          & 1 \% &  316 & toaster         & 6 \% &    1 \\
		76 & carrot          & 1 \% &    9 & carrot          & 2 \% &   12 \\
		77 & dining table    & 0 \% &   11 & dining table    & 1 \% &   33 \\
		78 & handbag         & 0 \% &    0 & handbag         & 1 \% &    9 \\
		79 & sports ball     & 0 \% &    0 & sports ball     & 1 \% &    1 \\
		80 & hair drier      & 0 \% &    0 & hair drier      & 0 \% &    0 \\
		\hline             
	\end{tabular}
	\caption[The degree of controllability by category for the distinct subset]{The degree of controllability by box category for the \textit{distinct} subset. The top and bottom ten box categories are listed regarding the percentage of matches. A match means that a box captions includes the category word (k@1) or one of the nearest neighbors (k@5). The arrows indicate the amount of position in an upwards or downwards direction, when switching between k@1 and k@5. There are in total 77,365 (k@1) or 52,107 (k@5) distinct box captions and 80 box categories. For category-wise total amounts see also Appendix \ref{AppendixC}.}
	\label{table:controlability_per_category_distinct}
\end{table}

\clearpage

\section{Discussion}

\paragraph{Research Question 1: Sensitivity towards spatial attention.} The first research question is asking, whether the asserted control on a caption generator's attention is causing the output to be different from what would have been otherwise produced.

Therefore, I let a trained image captioning model produce self-attending, control and box captions. These box captions, which were produced under the additional asserted control over the model's spatial attention, differed from the normally produced self-attending captions in up to 88.88 \% and differed from the control caption in up to 52.65 \%. These results indicate that the spatial attention manipulations are indeed causing the output captions to be different from self-attending captions in the majority of cases. 

These changes in the output captions are likely to be related to the manipulations in the model's spatial attention, because the same extracted image features and spatial attention vectors were used for all experiments. In addition, the words were sampled with the arg-max sampler that always produces the highest probable word at a certain time-step. Given this deterministic generation process, the experiments are expected to be reproducible.

Among the applied attention forcing methods, the unlimited step-wise fixed attention resulted in the highest method specific sensitivity. Given the results, this method describes an upper bound on the method sensitivity for the presented methods. The limited step-wise fixed attention shows an increasing effect on the method sensitivity with respect to the trained captioning model. Here, the results approach those of the unlimited method, when the fixation is performed over more iterations. Similar results are given for the additive attention. Here an increase in method sensitivity correlates with larger weights on the external attention. 

Nevertheless, the generator produces under the unlimited attention fixation still in 11.32\% of cases the same caption as the normally produced. These spatial attention vectors that induce no reacting to the captioning model might be derived from bounding boxes that refer to the main objects in an image. Then the caption generator is expected to produce the same caption that would have been normally produced. 
Other possible reasons are that the caption generator is biased towards easily detectable categories, which for example occur more often in the training split or that the bounding boxes were too small to put a recognizable constraint on the image. A deeper assessment of possible reasons will be beneficial to improve the results of this experiment, though I leave them for further work.

Furthermore, there are 47.35\% of box captions that are the same as the control caption for which the spatial attention vector is equally distributed over the whole image. A more detailed introspection of theses cases could reveal the same reasons as stated above. The image caption generator might have blind spots or an overall preference for specific categories. Then also a lower weighted signal in a certain image region might cause the generator to produce the according word.

\paragraph{Research Question 2: Controllability using spatial attention.} The second research question is asking, whether the spatial attention forcing methods are effective mechanisms to control the caption generation process in a predictable way.

Therefore, I applied three different interface methods using spatial attention vectors based on bounding boxes with annotated object categories. The fully trained image captioning model had to produce box captions under the expectation that the object category words are included in the resulting caption sequence. Given the results, such an expectation was full-filled in up to 28.56\% of the cases and in up to 58.17\% of the cases, where also words are allowed, which are under the learned word embedding similar to the category words in cosine-distance.
Thus the results indicate that a caption generation model with spatial attention is indeed controllable by the presented forcing methods. These expectations hold also true in 9.00\% of the cases on the distinct subset. Here box captions were discarded, when the image captioning model was already mentioning the object category in the normally produced self-attending caption. And again, allowing words similar to the object category raised the percentage up to 21.39\%.  

\clearpage

\paragraph{Category-wise controllability using spatial attention.} The best interface method to control the generator's spatial attention in a predictable way was the unlimited fixed attention technique. In reference to this method, I discuss here the possible reasons for the differences between exact and neighbor matching as well as the differences between the whole and the distinct set.

The most often mentioned category in relation to its count of appearances in images is cat. Here, the spatial attention is forced on cats in the according bounding boxes. In contrast to that, regarding the distinct subset, the cat category is only at position 15 with 32\%. This is a decrease in 14 positions between the evaluation sets. In the distinct subset, only 177 of the initially 1,542 bounding boxes are relevant for the evaluation. This reduction in 1,365 samples indicates that the model is biased towards cats in an image or that the images with cats contain in only rare cases further objects. A possible reason for the reduction of ``cat'' occurrences in the resulting box captions is that the model is already talking about the cat (or main object of the image) in obvious cases and in other cases the detection and therefore mentioning of the cat category gets harder. This would deny the working assumption that an attention aware captioning model automatically includes the dense captioning task. I leave such a further introspection for further work.


The difficulty for the trained captioning model to mention an object category for a certain spatial area might also rest upon the quality of the input images or on the capabilities of the learned vocabulary. This is in particular visible in the categorical results for the \textit{person} category. The person category has an increase in 52 positions from 75 to 23 in the distinct subset, when the nearest neighbors are also included for matching. We can intuitively guess that looking for ``person'' in the box caption is rather restrictive, when the caption might also state ``woman'' or ``man''. Therefore, there is a huge increase from 1\% to 32 \% in matches, when also ``man'', ``woman'', ``guy'' or ``someone'' is allowed to be included in the box caption for the distinct subset. The similar effect is visible for matches that incorporate the neighbors on the whole set.

\clearpage

On the downside, this simple choice of category alternatives based on the nearest neighbors results into rather uninformative increases of the category elephant regarding the whole set. Then also ``giraffe'', ``elephants'', ``bear'' or ``zebra'' are allowed to be included. Thus, the increasing position of the category elephant is here only grounded on the allowance of ``zebra'' in the resulting box caption. This is from a human perspective simply wrong, because a zebra is not an elephant. Therefore, a better evaluation would not include the nearest neighbors based on the cosine-distance, but rather use a manual chosen set of words for each of the eighty categories. Along with that, there could be a better handling of compound category words. For example, by determining the nearest neighbors from the added embedding vector of the two category words. In this work, I handle in the compound category words simply as two separate ones.

Furthermore, the categories are imbalanced towards a few. I use a general purpose dataset for object detection, because there are no datasets for this specific attention forcing task. On the one hand, this could show that general purpose datasets are usable to full-fill such a specific task. On the other hand, the dataset is highly biased towards the object category person with 37,051 of 117,798 annotated boxes (the second largest is chair with 5,337). Unfortunately, the person category is also the most diverse one with a lot of possible words with similar meaning. Ignoring the person category would increase the controllability measure to 16,71\% for k@1 and reduce it to 19.91\% for k@5.


I also leave for further work the more technical introspection of the effect of using different down-sampling algorithms to create the spatial attention vectors. Since now, I simply used the nearest interpolation to keep the box edges as original as possible and discard not large enough boxes. With other interpolation techniques also smaller boxes can be possibly examined. This technical fine-tuning would include to better understand the effect of the softmax applied on the resulting down-sampled spatial attention vectors. Here, the \textit{exact} attention values for specific image areas might be necessary to detect specific object categories. However, I hope this is not the case, because from this follow again inaccessible spatial attentions. In this work, the attention values slightly change with the box size for the according object.


\chapter{Experiment 2: Interface the Generator Using Another Model} 

\label{Chapter5} 

In this experiment I examined, if the output of a standard caption generator might be helpful to solve another task under the constraint that an external model provides spatial attention vectors as a guidance for the captioning network. Based on chapter \ref{Chapter4}, I chose the interface configurations with the best results among the methods. Then I manipulated the spatial attention of the trained caption generator from chapter \ref{Chapter3} to exemplary evaluate the relationship between the resulting image captions and a visual question answering task that was to be solved. 

As an exemplary producer for spatial attention vectors I chose the visual question answering model from \citet{lu_hierarchical_2016}. Then the image captioning model from \citet{xu_show_2015} had to generate what I call a \textit{word, phrase} or \textit{question caption} based on an according spatial attention vector. The generated captions were supposed to contain words which refer to either parts of the questions or answers. To provide insights regarding the effectiveness of this procedure, I introduced a measurement similar to the controllability from section \ref{sec:experiment1_setup}.

\clearpage

\section{Experimental Setup}

\subsection{A Visual Question Answering Model.} 

\begin{figure}
	\centering
	\includegraphics[width=.9\textwidth]{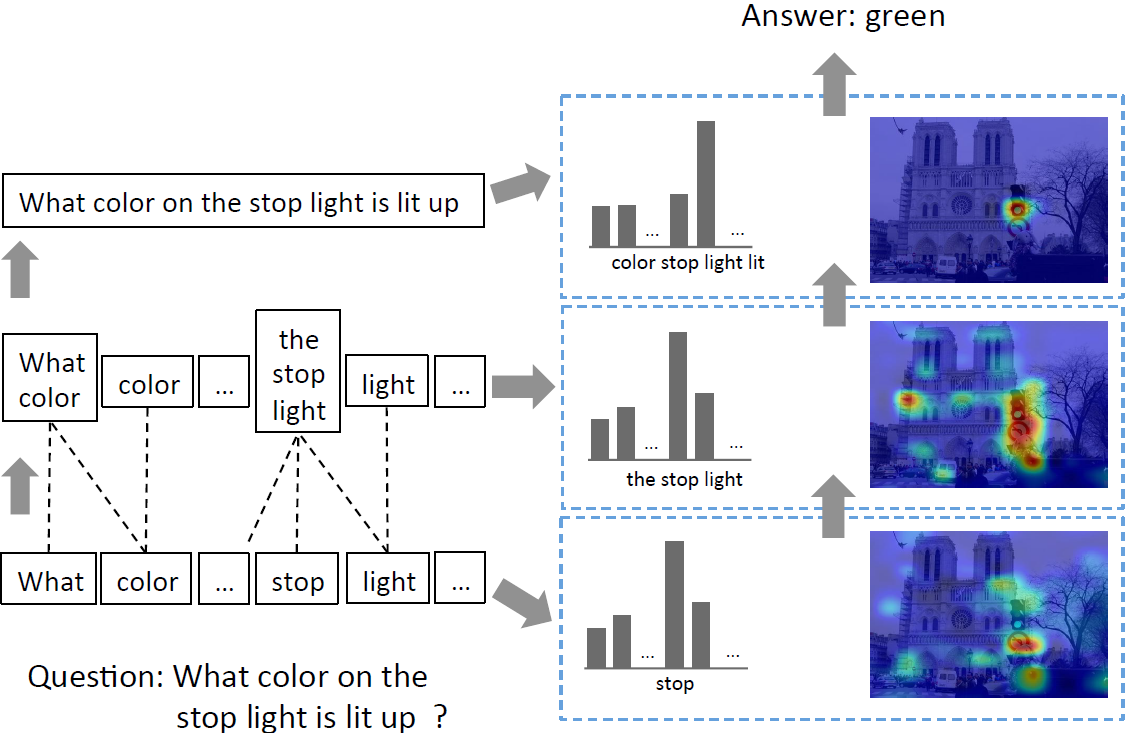}
	\caption[The conceptual idea of the hierarchical co-attention network]{The conceptual idea of the hierarchical co-attention network from \citet{lu_hierarchical_2016}. The question is decomposed into words and phrases or seen as a whole, on the left. The according attended regions are highlighted, on the right.}
	\label{fig:hicoatt_image}
\end{figure}

For this experiment, I reimplemented in Keras \parencite{francois_chollet_keras_2015} a hierarchical co-attention network for visual question answering \parencite{lu_hierarchical_2016}.\footnote{My source code is documented at \url{https://github.com/phisad/keras-hicoatt}} This network is similar to a captioning generator, but here the produced textual output is interpreted as the answer for a question. In such a way, the question acts as an additional context for the image that guides the spatial attention on the image. This idea had been extended to co-attention, which also predicts a language attention for words in the question based on the image. \citet{lu_hierarchical_2016} proposed a hierarchical approach in which they decompose the co-attention into a word, phrase and question level as shown in Figure \ref{fig:hicoatt_image}. These levels were supposed to represent different, but equally important aspects about the textual and visual input for answer prediction. 

\clearpage

Therefore \citet{lu_hierarchical_2016} defined a parallel co-attention network similar to the attention network of the captioning generator from \citet{xu_show_2015} where image and question attention are computed in parallel. The co-attention was based on a learned correlation matrix between image and question features 

\begin{equation}
	\text{C} = \text{tanh}( \text{Q}^{T} \textbf{W}_{c} \text{A} )
\end{equation}

with trainable parameters $\textbf{W}_{c} \in \mathbb{R}^{D \times D}$, the transpose question features $ \text{Q} \in \mathbb{R}^{D \times T}$ and the extracted image features $ \text{A} \in \mathbb{R}^{D \times L}$,  where \textit{D} is the number of word embedding dimensions or the number of feature maps and \textit{T} is the maximal question length while \textit{L} is the size of the feature map. 

This correlation matrix acted then as a trainable feature to predict the image attention (and question attention accordingly) by following

\begin{equation}
	\text{P}_{v} = \text{tanh}( \textbf{W}_{v} \text{A} + [\textbf{W}_{q} \text{Q}] \text{C})
\end{equation}

with trainable parameter $\textbf{W}_{v}, \textbf{W}_{q} $ and where $\text{P}_{v} \in \mathbb{R}^{D \times L}$ represents the image features conditioned on the question features. This allowed the network to adjust specific feature map signals based on the question features. Then the actual image attention was determined by summing up along the feature map dimension and applying the softmax

\begin{equation}
	\alpha_{v} = \text{softmax}( \sum_{j}^{D} [\text{P}_{v}(j,i)])
\end{equation}

so that the spatial attention was given exactly as from \citet{xu_show_2015} by $\alpha \in \mathbb{R}^{L}$, $\alpha \in (0,1)$ and $\sum{\alpha} = 1$. These spatial attentions were then applied to compute the context vectors likewise as in section \ref{sec:reimplementation}

\begin{equation}
\hat{v} = \sum_{j=1}^{L}(\alpha_{j} \text{A}(i,j))
\end{equation}

so that the image features are given by $\hat{v} \in \mathbb{R}^{D}$ and question features $\hat{q} \in \mathbb{R}^{D}$ accordingly (which are not explicitly shown here). Given these co-attention context vectors $\hat{q}_{\{w,p,s\}}$ for the question and $\hat{v}_{\{w,p,s\}}$ for the image at word, phrase and question level respectively, the answer was predicted using a feed-forward neural network as described by \citet{lu_hierarchical_2016} in the following

\begin{align}
	\text{h}_{w} =& \text{tanh}(\textbf{W}_{w} ( \hat{q}_{w} + \hat{v}_{w}) ) \\
	\text{h}_{p} =& \text{tanh}(\textbf{W}_{p} [(\hat{q}_{p} + \hat{v}_{p}) \oplus \text{h}_{w}] ) \\
	\text{h}_{s} =& \text{tanh}(\textbf{W}_{s} [(\hat{q}_{s} + \hat{v}_{s}) \oplus \text{h}_{p}] ) \\
	\text{y} =& \text{softmax}(\textbf{W}_{y} \text{h}_{s})
\end{align}

with the trainable parameters $\textbf{W}_{w}, \textbf{W}_{p}, \textbf{W}_{s}, \textbf{W}_{y}$ and $\oplus$ as the concatenation operation. In the end, the softmax output provided a probability distribution over most common answers in the dataset.

\subsection{Using Another Model as External Spatial Attention} 

As described in \citet{lu_hierarchical_2016}, I trained the hierarchical co-attention network for the  most common answers on the VQA 1.0 dataset \parencite{agrawal_vqa:_2017}. This dataset was in particular suitable, because the images were also taken from MSCOCO \parencite{lin_microsoft_2014} and thus the caption generator from chapter \ref{Chapter3} and the visual question answering model were trained on the same domain of images, although they solved different tasks.
The VQA 1.0 dataset provided 369,861 question-answer pairs for training (248,349) and validation (121,512) on the open-ended question answering task. In contrast to the multiple-choice task, where one of eighteen possible answers had to be selected, the open-ended answers were unconstrained. However, in the VQA 1.0 dataset, the average answer length was only $1.1 \pm 0.4 $ words. A reason for this might have been, that there were three sub-categories: \textit{Yes/No}, \textit{Numbers/Counting} and \textit{Other}, but only the sub-category Other was unrestricted in terms of word usage and represented only a share of 45,849 questions. 

\clearpage

\begin{table}
	\begin{tabular}{| r | r | r | r | r | r | r | r | r |}
		\hline
		Adpt.   & Tokens & Epoch &   Loss &     Acc.      & Yes/No      & Num.   & Other & Total \\
		\hline
		Yes 	  & Basic &    73 &   1.40 &  57.36 & \textbf{79.09} & \textbf{30.86} & \textbf{40.40} & \textbf{55.25} \\
		No 	      & Basic &    79 &   1.63 &  52.15 & 74.74 & 25.85 & 15.35 & 40.86 \\
		\hline
		Yes 	  & NLTK   &    68 &   \textbf{1.32} &  \textbf{59.73} & 77.42 & 21.06 & 12.53 & 40.09 \\
		No 	  	  & NLTK   &    61 &   1.69 &  51.17 & 72.21 & 32.34 & 10.96 & 38.41 \\
		\hline
		\hline
		\multicolumn{3}{|c}{\citet{lu_hierarchical_2016}} & \multicolumn{2}{c|}{parallel+VGG} & 79.50 & 38.70 & 48.30 & 60.10 \\
		\hline
	\end{tabular}
	\caption[The training scores for my hierarchical co-attention network]{The training scores for the re-implemented visual question answering network. The dropout rate was 50\% after each layer and 10\% after the top layer. Shown are the adapter (Adpt.) and tokenizer usage (Tokens) as well as the scores for the best epoch. Below, the reported scores from \citet{lu_hierarchical_2016}.}
	\label{table:vqa_scores}
\end{table}

\citet{lu_hierarchical_2016} joined the whole training and validation split for the final training to achieve better scores. Following this, I also joined the splits and determined the top 1,000 answers, which covered 84.43\% of all joint answers. 

Then, I prepared the questions and answers using the basic Keras tokenizer \parencite{francois_chollet_keras_2015} which split on whitespace and filtered all punctuation, plus tabs and line breaks, except the single upper quote. From the resulting tokenized question-answer pairs I removed those, that contained none of the top answers. As a result, the training split was reduced to 320,029 input pairs, which represented 86.53\% of the total joint split. Given these remaining question-answering pairs, I built a vocabulary resulting into 20,946 known tokens and a maximal question length of 22 words. As image features I used the same extracted ones as in chapter \ref{Chapter4}. 

I trained two different configurations: one using an adaption layer for the image features with a dropout rate of 10\% and another without an image feature adapter. The idea of introducing an adaption layer was motivated by the fact, that the VGG-19 image feature extractor had been pre-trained on the ImageNet dataset \parencite{deng_imagenet:_2009} which represents another domain.

I performed the same steps using the NLTK tokenizer \parencite{loper_nltk:_2002} as an alternative tokenization technique. As a result, the training split was reduced to 320,029 input pairs, which were 86.53\% of the total joint split. From these remaining question-answering pairs I built a vocabulary which consisted of 21,747 known tokens and a maximal question length of 23 words. 

\clearpage

\begin{figure}
	\centering
	\subfloat[Question: What is the dog doing? Answer: sleeping]{
		\includegraphics[width=\textwidth]{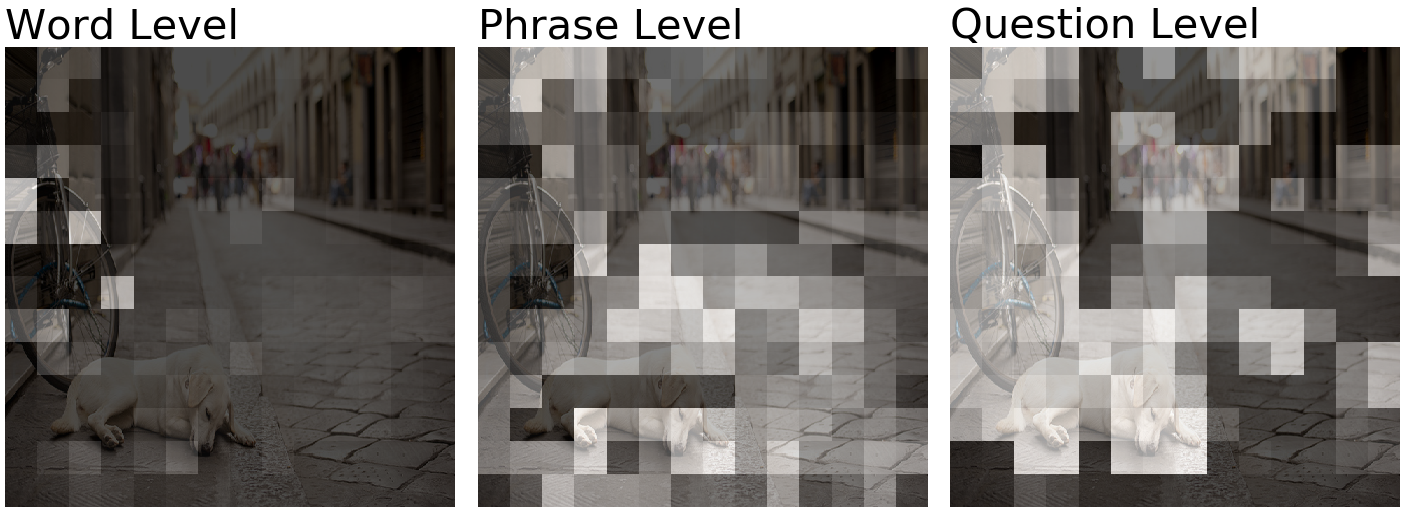}
	}
	\hfill
	\subfloat[Question: Where is the dog laying? Answer: sidewalk]{
		\includegraphics[width=\textwidth]{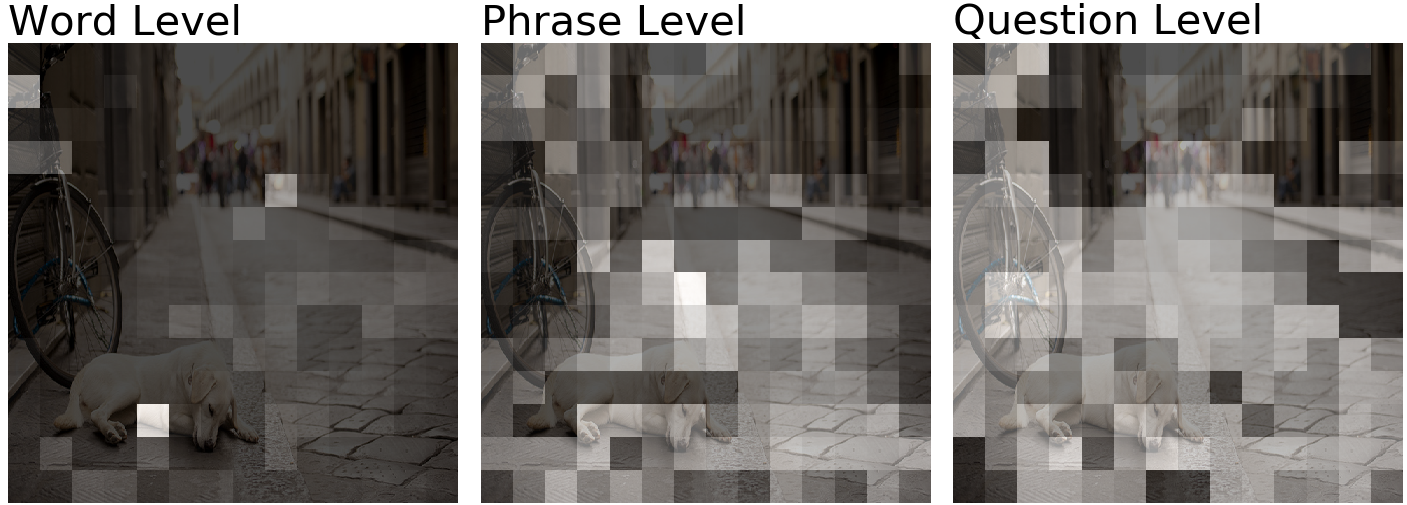}
	}
	\caption[A word, phrase and question level attention]{The spatial attention at word, phrase and question level of the hierarchical co-attention network regarding the specific questions about the sample image given below.}
	\label{fig:vqa_dog_attention}
\end{figure}

As shown Table \ref{table:vqa_scores}, the NLTK tokenized dataset lead to worse scores than with basic tokenization, although a better training accuracy and a lower loss were achieved. Anyhow, I simply chose the model with the highest scores, because the model still achieves a baseline score and was able to produce the necessary spatial word, phrase and question attention for my experiment. 

The best model with basic tokenization achieved 55.25 \% in total. Based on this model, I was not able to reproduce the scores reported by \citet{lu_hierarchical_2016} with a difference of $-0.41\%$ (Yes/No), $-7.84\%$ (Counting), $-7.90\%$ (Other) and  $-4.85\%$ in total. I leave possible explanations for further work.

\clearpage

Then, I extracted the spatial attention vectors for the word, phrase and question level as shown in Figure \ref{fig:vqa_dog_attention}. I performed the extraction only for the 45,849 questions of type \textit{Other}, because the caption generator was intended to produce a contextual description or even a long answer instead of just ``yes'', ``no'' or a single number. Thus, only 30,684 images were involved during the experiment.

Finally, I manipulated the spatial attention of the trained image captioning network from section \ref{sec:shatt} with the external spatial attention vectors extracted from the hierarchical co-attention model. For this, I was using the interface configurations with the best results in chapter \ref{Chapter4} in the following ways: 

\begin{itemize}
	\item I fed the external spatial attention vector at each time step to caption generator for the whole generation process while dismissing the caption generator's own attention. 
	\item I fed the external spatial attention vector to caption generator for the \textit{first six time-steps} while dismissing the caption generator's own attention. After this, the caption generator is again using its own attention.
	\item I added the the external spatial attention vector to the one predicted by the caption generator's attention model with a \textit{weighting factor of three}.
\end{itemize}

\subsection{Statistics}

\paragraph{Result Captions.} The trained image captioning model had to produce a caption for each level of the hierarchical co-attention network, namely a \textit{word, phrase} and \textit{question caption}. Moreover, I reused the already produced \textit{control captions} to set the resulting captions in relation with uniform spatial attention. In addition, I included the \textit{self-attending captions} from the first experiment to measure the general performance of the captioning model. 

Unfortunately, there were collectively 2,071 images for which not a single bounding box and thus no control caption was given. This further reduced the amount of relevant \textit{Other} questions from 45,849 to 42,871. Nonetheless, I generated five captions per question-answer pair for the remaining images, which lead altogether to 214,355 level, control and self-attending captions.

\clearpage

\paragraph{Degree of Usefulness.} The expectation was that the resulting word, phrase and question captions describe details about the question or answer of the visual question answering task. Thus, I suggested the \textit{degree of usefulness} for an interface method as a measurement by checking that the resulting caption included words of the question or answers. For this, I determined the stemmed set of words for the question, answer and caption respectively by using the Porter stemmer \parencite{porter_readings_1997}. In addition, I removed stop words from these sets. Then I counted each caption stemmed word as matching, when it was included in the question, in the answer or in one of both. 

\section{Results}

\paragraph{Qualitative Results.} Here, I give exemplary results for the image shown in Figure \ref{fig:vqa_dog_attention} to provide a better understanding of the quantitative evaluation. 

The trained caption generator produced under the additive attention with a weighting factor of three and fed with the spatial attention extracted level-wise for the question \textit{What is the dog doing?} (Answer: sleeping), the captions:

\begin{itemize}
	\setlength\itemsep{-0.7em}
	\item 	\textbf{word caption\hspace{.6cm}:} a bicycle is parked next to a bike
	\item 	\textbf{phrase caption\hspace{.4cm}:} a dog is sitting on the ground with a bicycle
	\item 	\textbf{question caption\hspace{.0cm}:} a dog is sitting on a leash on a bike
	\item 	\textbf{control caption\hspace{.3cm}:} a dog is sitting on the sidewalk next to a bike
	\item 	\textbf{self-attending\hspace{.5cm}:} a dog is sitting on the sidewalk next to a bike
\end{itemize}

First the question, answer and captions were tokenized and converted to a stemmed set of words excluding stop words, so that the question set is \{dog\}, the answer set is \{sleep\} and the according caption sets were 

\begin{itemize}
	\setlength\itemsep{-0.7em}
	\item 	\textbf{word caption\hspace{.6cm}:} \{bicycl bike next park\}
	\item 	\textbf{phrase caption\hspace{.4cm}:} \{bicycl dog ground sit\}
	\item 	\textbf{question caption\hspace{.0cm}:} \{bike dog leash sit\}
	\item 	\textbf{control caption\hspace{.3cm}:} \{bike dog next sidewalk sit\}
	\item 	\textbf{self-attending\hspace{.5cm}:} \{bike dog next sidewalk sit\}
\end{itemize}

Given these resulting caption sets, the degree of usefulness in reference to the answer is zero, because none of the captions included \textit{sleep}. However, all but the word caption set were including the word \textit{dog}. Therefore the phrase, question, control and self-attending caption were counted into the usefulness measure as a positive sample regarding the given question.


\begin{table}
	
	\begin{tabular}{| r | r | r | r || r | r || r | r |}
		\hline
		\multicolumn{2}{|c|}{Attention}  & \multicolumn{6}{c|}{Caption contains words that are also included} \\
		Forcing   &  Level     & \multicolumn{2}{c|}{in answer } 
		& \multicolumn{2}{c|}{in question}
		& \multicolumn{2}{c|}{in one of both} \\
		\hline
		&       word & 26.33 \% & 11,287 & 37.99 \% & 16,285 & 54.77 \% & 23,479 \\
		unlimited &     phrase & 19.79 \% &  8,483 & 33.38 \% & 14,309 & 46.25 \% & 19,827 \\
		&   question & 22.08 \% &  9,466 & 35.71 \% & 15,310 & 49.90 \% & 21,391 \\
		\hline
		&       word & 25.51 \% & 10,936 & 37.25 \% & 15,971 & 53.56 \% & 22,960 \\
		limited-6 &     phrase & 20.44 \% &  8,761 & 34.07 \% & 14,607 & 47.20 \% & 20,233 \\
		&   question & 21.74 \% &  9,321 & 35.35 \% & 15,155 & 49.35 \% & 21,157 \\
		\hline
		&       word & \textbf{26.45} \% & \textbf{11,340} & 38.96 \% & 16,703 & \textbf{55.20} \% & \textbf{23,664} \\
		additive-3 &     phrase & 20.69 \% &  8,868 & 34.71 \% & 14,880 & 47.74 \% & 20,466 \\
		&   question & 22.47 \% &  9,634 & 36.27 \% & 15,550 & 50.32 \% & 21,572 \\
		\hline
		\multicolumn{2}{|c|}{control} & 21.12 \% &  9,055 & 36.56 \% & 15,673 & 49.85 \% & 21,372 \\
		\multicolumn{2}{|c|}{self-attending} & 23.90 \% & 10,248 & \textbf{39.65} \% & \textbf{16,997} & 53.50 \% & 22,934 \\
		\hline
	\end{tabular}
	\caption[The degree of usefulness of generated captions per level]{The degree of usefulness for generated captions grouped by interface method and co-attention level. The results are given in absolute counts and as the percentage of 42,871 questions from the sub-category \textit{Other} in the VQA 1.0 dataset.}
	\label{table:vqa_results}
\end{table}

\paragraph{Quantitative Results.}


As shown in Table~\ref{table:vqa_results}, the trained captioning model produced in 55.20\% of the cases a question or answer related caption for the extracted word level attention, when using the step-wise additive method with a weighting factor of three. This was the highest measured degree in this experiment, which could be decomposed in 26.45\% of the captions including answer words and 38.96\% containing question words. Only, the normally produced self-attending captions included in additionally 0.64\% of the cases words of the questions. Hence, this was the highest for this segment. The smallest degree of method-wise usefulness was given for the result captions of the unlimited step-wise fixed attention using the extracted phrase level attentions. Here, in least of the cases a caption word was included in the answer (19.79\%) or in the question (33.38\%) respectively. This also resulted into the lowest overall usefulness with only 46.25\% of the cases in which a caption word is either contained in the answer or in the question. 

\clearpage

The best captions were produced from the caption generator, when fed with word level spatial attention vectors. This achieved the highest percentage over all segments. The second best were the question level attentions and the least useful result captions were based on the phrase level attention vectors.

\section{Discussion}

\paragraph{Research Question 3.} The last research question is asking, whether spatial attention is an useful interface for image captioning models to integrate their capabilities into tasks like visual question answering.

Therefore I re-implemented the hierarchical co-attention network from \citet{lu_hierarchical_2016} and trained it an the VQA 1.0 dataset \parencite{agrawal_vqa:_2017}. The trained network was not achieving the reported scores from \citet{lu_hierarchical_2016}, but was still modifiable for extraction of word, phrase and question level spatial attention vectors. I extracted these for all questions about images of the open-ended \textit{Other} sub-task. Then, I let the caption generator from section \ref{sec:shatt} produce word, phrase and question captions. In addition, I added the control and self-attending captions produced in the first experiment. 
Finally, I evaluated the resulting captions by quantifying how many stemmed caption words were included in the likewise stemmed question or answer, while ignoring stop words. Given the results, the caption generator produced in up to 55.20\% of the cases captions that contained words about the question or answer. This measurement indicates that the caption generator is indeed capable to produce useful information for almost every second question-answer pair, when fed with external spatial attention using the interface methods.

The highest results were given for word level spatial attentions. This is reasonable as these are also the ones with the highest object focus as exemplary shown in Figure \ref{fig:vqa_dog_attention}. Similar to the image captioning model, the word attention from the hierarchical co-attention network represents an alignment of specific objects in the image with individual question words. Therefore, the word attention is a good guidance for the caption generator to tell something useful about the image.

Furthermore, the human annotators are likely to provide a question that includes the specific object names, because there are on average only 3.5 different object categories \parencite{lin_microsoft_2014}. In contrast to that the phrase and question attention is rather distributed over the whole image for the sample image. As such, these spatial attentions are functioning more like the control caption, where the attention is distributed uniformly over the image. This would explain, why the results for captions based on the word level attentions are closer to the results from self-attending captions as shown in Table \ref{table:vqa_results}, but phrase and question level attention captions result into scores more similar to the control caption ones. 

Interestingly enough, the self-attending captions provided in more cases question related clues than the best interface method. This result might be highly dataset dependent. The MSCOCO dataset \parencite{lin_microsoft_2014} has often only a few main objects within the image. Thus, the human annotators are likely to include these into the question. Then also a general image caption is capable to describe the overall scene. Therefore, the self-attending captions might be taken as the baseline for a specific dataset. 

Seeing the normally produced captions as a baseline, all interface methods were less informative about questions. Nonetheless, the interface methods performed better than the baseline with respect to answers, when they were conditioned by the word level attention. A reason for this might be that the external spatial attention is indeed concentrated within bounding boxes of objects which are relevant for answering. This connection could be automatically examined for question-answer pairs, when the answer is represented by a single object in the image. In such a case, the bounding box could be compared with the word attention produced by the answering network. I leave such an assessment for further work. 

In addition, it would be interesting to use a dataset with more object categories like the \textit{ReferItGame} dataset \parencite{kazemzadeh_referitgame:_2014} to create referring expressions about objects in the image. Furthermore, the even larger \textit{Visual Genome} dataset \parencite{krishna_visual_2016} with both more object categories and questions per image could further improve the interface methods.
 

\chapter{Conclusion and Further Work} 

\label{Chapter6} 


The modern deep learning architectures for visual question answering are able to provide a short answer for a question about an image. Nevertheless, the internal workings in terms of decision making stay often unclear to an external observer. Since some of these model are showing a varying spatial attention on specific image regions during the conclusion process, the idea of this work was to translate these spatial attentions into natural language to provide a simpler and more expressive access to the model's function.

A problem with this idea was, that a supervised training of such models required a dataset for which the spatial attention under a textual output is known. To my knowledge, datasets for this specific problem do not exist, because they are hard and expensive to create, even though, there are attempts to synthetically produce datasets with text-attention conditionals for supervised training \parencite{cornia_show_2018}.

Therefore, instead of creating such a specific dataset, I took a known image captioning along with a known visual question answering architecture which both rely on image attention and assumed a possible connection between them. The benefit of this approach was that the networks can be trained independently and that relevant datasets already exist. After training, the caption generator was supposed to translate the spatial attention dynamics of the visual question answering model during the conclusion process. Here, the spatial attention is seen as an external interface to the caption generator.

\clearpage

First, I evaluated, if the caption generator is actually reacting to modification in the spatial attention, when it is externally applied in three different ways: fixed over the whole caption generation, fixed for the first time-steps and added to the caption generator's attention. The experimental results have shown that the captioning model is reacting to method dependent changes in up to 52.65\% of the cases. This means that the result caption is different from what would have been otherwise produced in at least a single word under the assumption that the generation process with maximum sampling is deterministic and reproducible.

Secondly, the captions produced with the interface methods were assumed to be predictable. This would indicate that the interface establishes control over the generation process. To test this, I derived spatial attention vectors from annotated bounding boxes of a standard object detection dataset with eighty object categories. In this way, the bounding box provide a label for an image region into which I forced the spatial attention using the methods from above. The generated captions include in 9.00\% of the cases categories which are unmentioned in normally generated captions. In such a sense, the image caption generator is controllable to some degree by using externally provided spatial attention, when this is focusing on specific parts of the image.


After I tested the caption generator's sensitivity and controllability towards external spatial attention, I established such a link between two separate and already trained models. Therefore, I extracted the word, phrase and question level spatial attention vectors from a hierarchical co-attention network for visual question answering while the model was producing answers for questions in a standard VQA dataset. These spatial attention vectors were then applied on the caption generator during the generation process using the methods from above. The captions produced under word level attention included words of the answer or question in 55.20\% of the cases. The other way around, in about half of the cases, a visual question answering network could have included information generated by a separate captioning model using spatial attention as an external interface. However, this represents only a slight margin over the baseline for which the normally generated captions also included words related to 53.50\% of the question-answer pairs.

\clearpage


This work indicates that spatial attention as an interface for image captioning models is a useful method to assert external control over its generation process. In this way, another model's decision making which is based on such an attention mechanism is expressable in a more human accessible way. A great benefit is that the usage as an external interface enables a separated training and allows to make use of various existing standard datasets. The image captioning model acts as an arbitrary complex spatial attention to natural language converter and is simply attachable as an external knowledge base. Possibly, any other model that has a decision making process which involves spatial attention can make use of the caption generator's interface.

Nevertheless, there still exist problems with the proposed approach of this thesis. The assumption that a caption generator inherently includes the dense captioning task was necessary, because specific datasets with caption-under-attention conditionals do not exist. It remains unclear, how strong this assumption really is. In addition, the used MSCOCO dataset is rather restrictive and unbalanced in terms of categories \parencite{lin_microsoft_2014}. Future work will include the usage of a cleaner and more balanced datasets like Visual Genome \parencite{krishna_visual_2016} for the proposed evaluation tasks. Given such datasets, the captioning model's performance is expected to improve. 

Moreover, I was unable to produce the reported scores for the hierarchical co-attention network. This missing performance could have had a significant impact on the results for the degree of usefulness regarding word, phrase and question level captions. Further work includes the usage of other better performing models and the extension to other tasks than VQA. 
Last but not least, there is a proportion of cases in which the caption generator shows no reaction to the proposed interface methods. I leave an in detail introspection of these cases for further work. Such an assessment will help to improve and better contextualize the results of this thesis.

In the end, I think that modifying the spatial attention of a standard neural image captioning model introduces an interesting new research direction for natural language generation, which will allow researchers to handle and understand the complexities of these models more easily.


\appendix 



\chapter{Figures of Regularizer Effects} 

\label{AppendixA} 

{
	\centering
	\includegraphics[width=.8\textwidth]{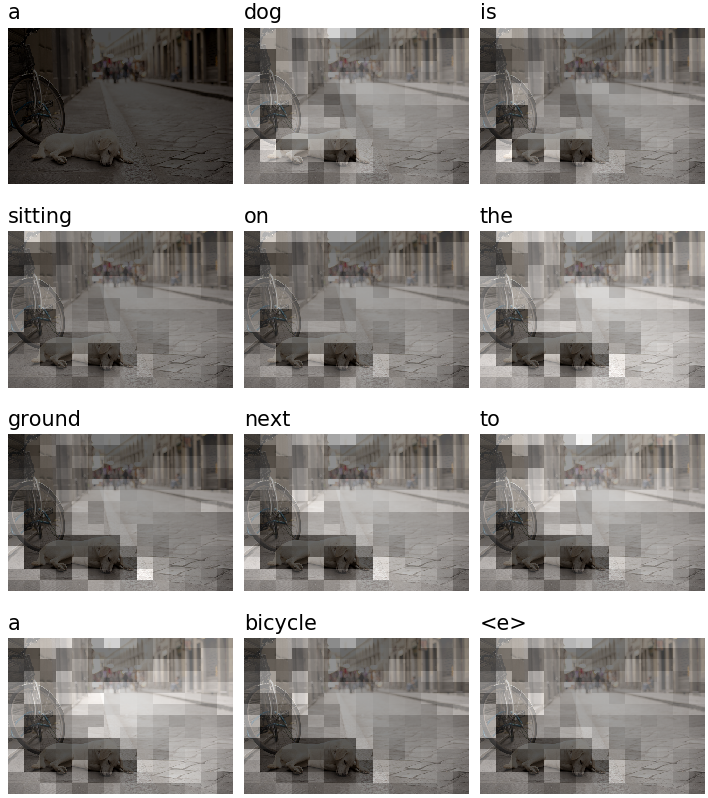}
	\captionof{figure}[Spatial attention of a regularized model with $\lambda = 0.001$]{The spatial attention at each time-step for the reimplemented \textit{Show, Attend and Tell} network with $\lambda = 0.001$
		\label{fig:dog_alternating_seq_01}}
}

\begin{figure}[p]
	\centering
	\includegraphics[width=.8\textwidth]{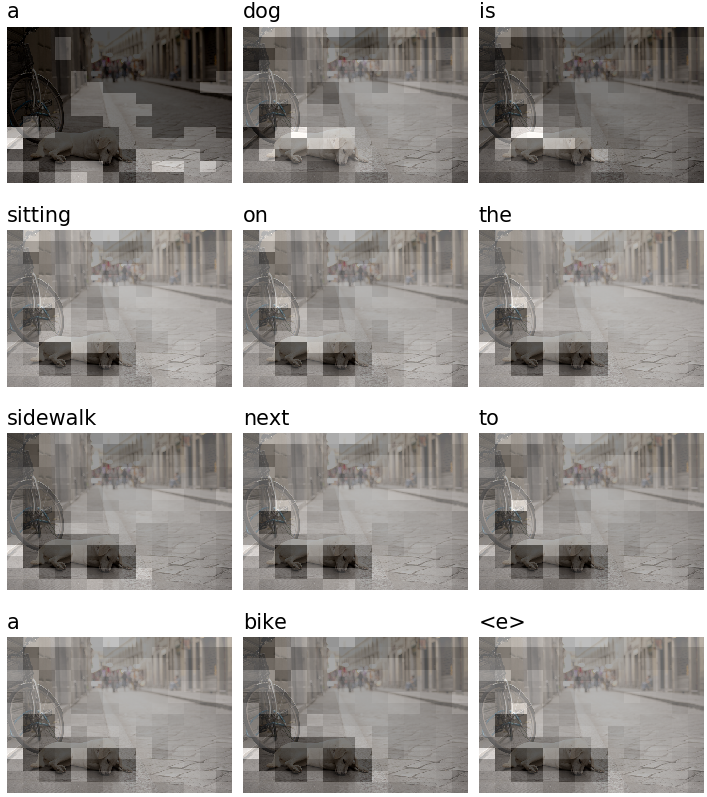}
	\captionof{figure}[Spatial attention of a regularized model with $\lambda = 0.005$]{The spatial attention at each time-step for the reimplemented \textit{Show, Attend and Tell} network with $\lambda = 0.005$
		\label{fig:dog_alternating_seq}}
\end{figure}
	
\begin{figure}[p]
	\centering
	\includegraphics[width=.8\textwidth]{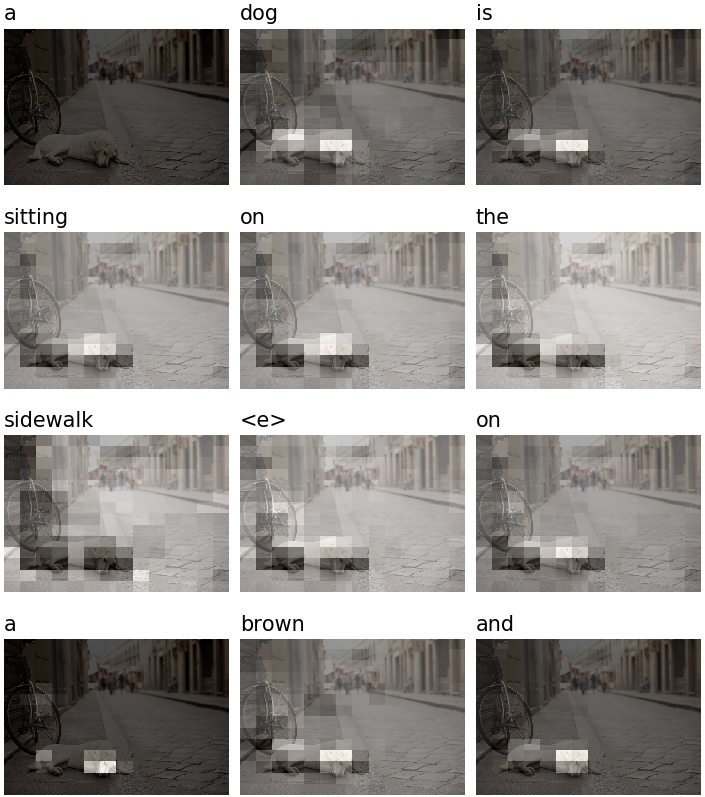}
	\captionof{figure}[Spatial attention of a regularized model with $\lambda = 0.010$]{The spatial attention at each time-step for the reimplemented \textit{Show, Attend and Tell} network with $\lambda = 0.010$
		\label{fig:dog_alternating_seq_10}}
\end{figure}


\chapter{Figures of Interface Methods} 

\label{AppendixB} 

{
	\centering
	\includegraphics[width=\textwidth]{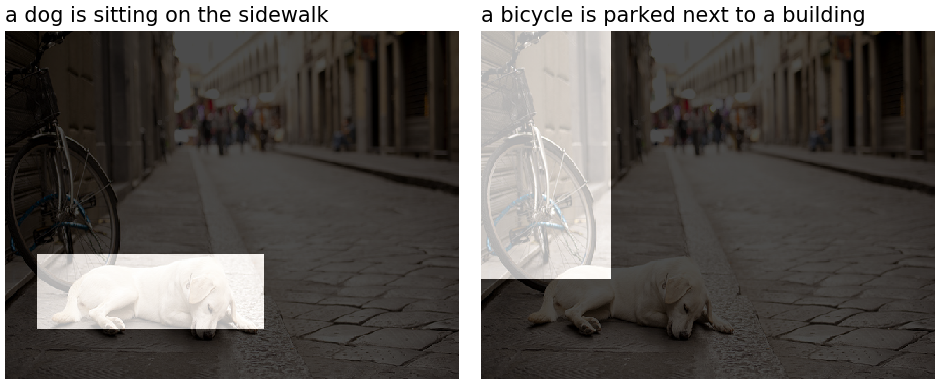}
	\captionof{figure}[Spatial attention under the unlimited step-wise fixed attention]{The spatial attention under the unlimited step-wise fixed attention.%
		\label{fig:dog_fixed_sum}}
}

\begin{figure}[p]
	\centering
	\includegraphics[width=\textwidth]{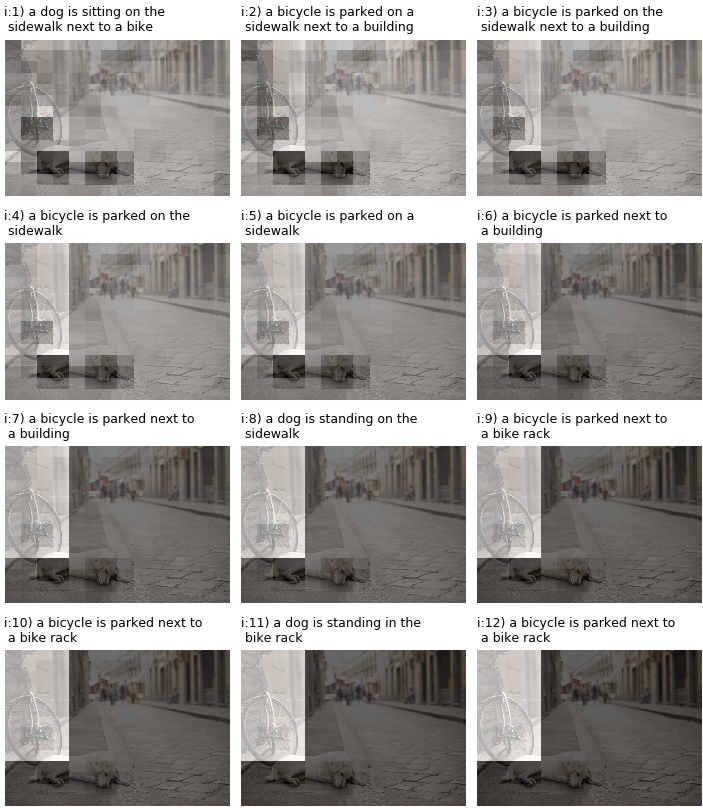}
	\captionof{figure}[Spatial attention under the limited step-wise fixed attention]{The spatial attention under the limited step-wise fixed attention with different configuration for the fixation length. Here the fixation length ranges from $[1-12]$. This means a fixation up to the twelfth time-step.%
		\label{fig:dog_semi_fixed_sum_multiple_iterations_on_bicycle}}
\end{figure}

\begin{figure}[p]
	\centering
	\includegraphics[width=\textwidth]{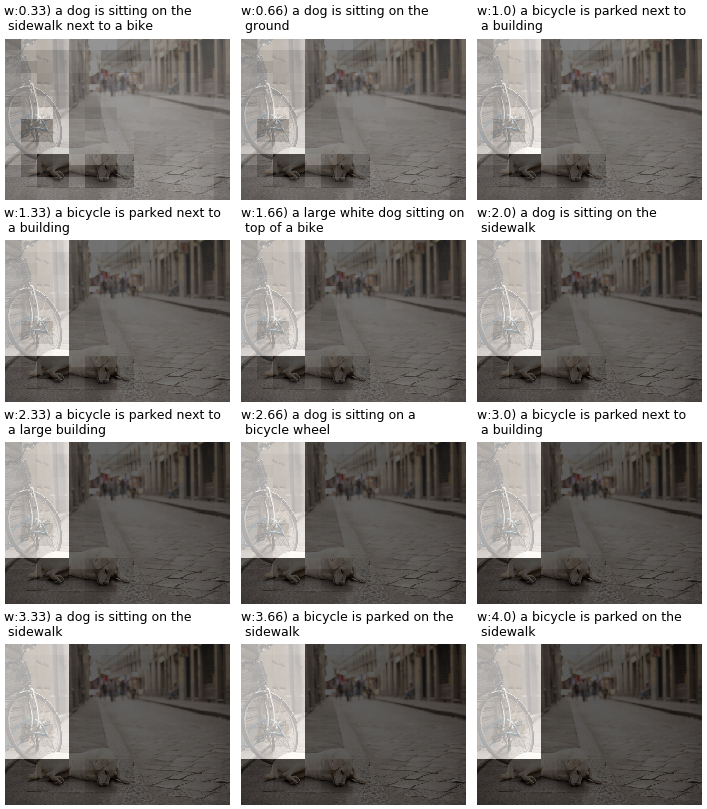}
	\captionof{figure}[Spatial attention under the step-wise additive attention]{The summed spatial attention under the step-wise additive attention with differently chosen weights. The weights range from $[0-4]$. This means that the externally provided attention has up to four times more weight than the attention predicted by the generator.%
		\label{fig:dog_dynamic_sum_multiple_weights_on_bicycle}}
\end{figure}

The spatial attention for the different interface methods shown in Figures \ref{fig:dog_fixed_sum}, \ref{fig:dog_semi_fixed_sum_multiple_iterations_on_bicycle} and \ref{fig:dog_dynamic_sum_multiple_weights_on_bicycle}. The attention is summed up over all time-steps during the caption generation. This allows to present the image along with the spatial attention at once.

The images show the attention for the bounding boxes of a dog or a bicycle. Other bounding boxes for this picture were discarded, because they were not larger than the median. 

Above the image, the produced caption is shown according to the attention. The caption generation was stopped at the end word, although there might be produced additional words until the maximal caption length is reached.

\chapter{Result Tables for Categorical Controllability} 

\label{AppendixC} 

In the following I present the categorical controllability results for the top and bottom ten regarding the whole and distinct set. The Tables \ref{table:results_categorical_whole_k1} and \ref{table:results_categorical_whole_k5} show results for the whole evaluation set. The Tables \ref{table:results_categorical_distinct_k1} and \ref{table:results_categorical_distinct_k5} show results for the distinct subset.

For the distinct subset, the resulting box caption which were produced under bounding boxes that contain already mentioned objects. In these cases the trained caption generator was talking about the box object anyway.

I only show the results for the upper ten and lower ten positions, because these are the most interesting ones and here is not enough space to list all eighty categories. The intermediate positions were only included in Chapter \ref{Chapter4} to express the position shifts were.

The purpose for this appendix is to document also the total amount of samples according to a category. The results presented in Chapter \ref{Chapter4} were only showing the relative and absolute results for the matches. Furthermore, the object category identifiers are provided.

\begin{table}[p]
	\centering
	\begin{tabular}{| r | r | l | r | r | r | r |}
		\multicolumn{7}{ c }{Box Captions including the Category Word} \\
		\hline
		Pos. & Id. & Category Name & Matches & & Total &  \\
		\hline
		1 & 17 & cat             & 1370 & 89 \% &  1542 & 1 \% \\
		2 &  7 & train           & 1209 & 85 \% &  1426 & 1 \% \\
		3 & 59 & pizza           & 1205 & 85 \% &  1414 & 1 \% \\
		4 & 70 & toilet          & 1067 & 85 \% &  1250 & 1 \% \\
		5 & 85 & clock           &  628 & 85 \% &   741 & 1 \% \\
		6 & 11 & fire hydrant    &  332 & 83 \% &   399 & 0 \% \\
		7 & 24 & zebra           & 1094 & 77 \% &  1412 & 1 \% \\
		8 & 23 & bear            &  323 & 77 \% &   419 & 0 \% \\
		9 & 18 & dog             & 1175 & 76 \% &  1536 & 1 \% \\
		10 & 20 & sheep           &  950 & 76 \% &  1256 & 1 \% \\
		\hline
		\hline
		71 & 53 & apple           &   41 & 7 \% &   620 & 1 \% \\
		72 & 27 & backpack        &   55 & 6 \% &   883 & 1 \% \\
		73 & 80 & toaster         &    1 & 4 \% &    23 & 0 \% \\
		74 &  1 & person          &  877 & 2 \% & 37051 & 31 \% \\
		75 & 64 & potted plant    &   22 & 2 \% &  1415 & 1 \% \\
		76 & 57 & carrot          &   19 & 2 \% &   808 & 1 \% \\
		77 & 67 & dining table    &   12 & 0 \% &  3911 & 3 \% \\
		78 & 31 & handbag         &    0 & 0 \% &  1026 & 1 \% \\
		79 & 37 & sports ball     &    0 & 0 \% &   119 & 0 \% \\
		80 & 89 & hair drier      &    0 & 0 \% &    39 & 0 \% \\
		\hline             
	\end{tabular}
	\caption[Results for the categorical controllability]{The top and bottom ten results for the categorical controllability in relation to all 117,798 box captions. The matches are related to the total amount of samples for the \\ category.}
	\label{table:results_categorical_whole_k1}
\end{table}

\begin{table}[p]
	\centering
	\begin{tabular}{| r | r | l | r | r | r | r |}
		\multicolumn{7}{ c }{Box Captions including the Category Word or } \\
		\multicolumn{7}{ c }{one of the 5 Nearest Words} \\
		\hline
		Pos. & Id. & Category Name & Matches & & Total &  \\
		\hline
		 1 & 24 & zebra           & 1386 & 98 \% &  1412 & 1 \% \\
		2 & 17 & cat             & 1445 & 94 \% &  1542 & 1 \% \\
		3 & 25 & giraffe         & 1409 & 94 \% &  1493 & 1 \% \\
		4 &  7 & train           & 1264 & 89 \% &  1426 & 1 \% \\
		5 & 22 & elephant        & 1169 & 87 \% &  1343 & 1 \% \\
		6 & 59 & pizza           & 1221 & 86 \% &  1414 & 1 \% \\
		7 & 23 & bear            &  362 & 86 \% &   419 & 0 \% \\
		8 & 70 & toilet          & 1067 & 85 \% &  1250 & 1 \% \\
		9 & 85 & clock           &  630 & 85 \% &   741 & 1 \% \\
		10 & 11 & fire hydrant    &  339 & 85 \% &   399 & 0 \% \\
		\hline
		\hline
		71 & 53 & apple           &   88 & 14 \% &   620 & 1 \% \\
		72 & 55 & orange          &   96 & 11 \% &   900 & 1 \% \\
		73 & 40 & baseball glove  &   14 & 11 \% &   126 & 0 \% \\
		74 & 27 & backpack        &   71 & 8 \% &   883 & 1 \% \\
		75 & 67 & dining table    &  190 & 5 \% &  3911 & 3 \% \\
		76 & 80 & toaster         &    1 & 4 \% &    23 & 0 \% \\
		77 & 57 & carrot          &   22 & 3 \% &   808 & 1 \% \\
		78 & 37 & sports ball     &    3 & 3 \% &   119 & 0 \% \\
		79 & 31 & handbag         &   20 & 2 \% &  1026 & 1 \% \\
		80 & 89 & hair drier      &    0 & 0 \% &    39 & 0 \% \\
		\hline             
	\end{tabular}
	\caption[Results for the categorical controllability with neighbors]{The top and bottom ten results for the categorical controllability in relation to all 117,798 box captions. The matches allow here also to include the nearest neighbor words. The matches are related to the total amount of samples for the category.}
	\label{table:results_categorical_whole_k5}
\end{table}
	
\begin{table}[p]
	\centering
	\begin{tabular}{| r | r | l | r | r | r | r |}
		\multicolumn{7}{ c }{Box Captions including the Category Word} \\
		\multicolumn{7}{ c }{(on the Distinct Subset) } \\
		\hline
		Pos. & Id. & Category Name & Matches & & Relevant &  \\
		\hline
		1 & 24 & zebra           &  275 & 67 \% &   413 & 0 \% \\
		2 & 59 & pizza           &  113 & 48 \% &   233 & 0 \% \\
		3 &  4 & motorcycle      &  356 & 46 \% &   768 & 1 \% \\
		4 & 25 & giraffe         &  227 & 45 \% &   499 & 0 \% \\
		5 & 14 & parking meter   &   23 & 44 \% &    52 & 0 \% \\
		6 & 73 & laptop          &  193 & 43 \% &   448 & 0 \% \\
		7 & 70 & toilet          &   74 & 42 \% &   176 & 0 \% \\
		8 & 11 & fire hydrant    &   39 & 42 \% &    92 & 0 \% \\
		9 & 85 & clock           &   51 & 38 \% &   134 & 0 \% \\
		10 & 41 & skateboard      &   69 & 36 \% &   193 & 0 \% \\
		\hline
		\hline
		71 & 27 & backpack        &   46 & 6 \% &   771 & 1 \% \\
		72 & 80 & toaster         &    1 & 6 \% &    18 & 0 \% \\
		73 & 50 & spoon           &   19 & 5 \% &   380 & 0 \% \\
		74 & 64 & potted plant    &   20 & 2 \% &  1196 & 1 \% \\
		75 &  1 & person          &  316 & 1 \% & 32673 & 28 \% \\
		76 & 57 & carrot          &    9 & 1 \% &   631 & 1 \% \\
		77 & 67 & dining table    &   11 & 0 \% &  3318 & 3 \% \\
		78 & 31 & handbag         &    0 & 0 \% &   912 & 1 \% \\
		79 & 37 & sports ball     &    0 & 0 \% &   109 & 0 \% \\
		80 & 89 & hair drier      &    0 & 0 \% &    37 & 0 \% \\
		\hline             
	\end{tabular}
	\caption[Results for the distinct categorical controllability]{The top and bottom ten results for the categorical controllability in relation to the 77,365 box captions in the distinct subset. Box captions were ignored, when their category was already mentioned by the normal caption. The matches are related to the total amount of samples for the category in the distinct subset.}
	\label{table:results_categorical_distinct_k1}
\end{table}

\begin{table}[p]
	\centering
	\begin{tabular}{| r | r | l | r | r | r | r |}
		\multicolumn{7}{ c }{Box Captions including the Category Word or } \\
		\multicolumn{7}{ c }{one of the 5 Nearest Words (on the Distinct Subset) } \\
		\hline
		Pos. & Id. & Category Name & Matches & & Relevant &  \\
		\hline
		1 & 24 & zebra           &   28 & 68 \% &    41 & 0 \% \\
		2 & 43 & tennis racket   &  198 & 56 \% &   355 & 0 \% \\
		3 &  4 & motorcycle      &  374 & 54 \% &   696 & 1 \% \\
		4 &  5 & airplane        &  184 & 54 \% &   338 & 0 \% \\
		5 & 59 & pizza           &  114 & 52 \% &   218 & 0 \% \\
		6 & 11 & fire hydrant    &   34 & 44 \% &    77 & 0 \% \\
		7 & 70 & toilet          &   74 & 42 \% &   176 & 0 \% \\
		8 & 73 & laptop          &  154 & 41 \% &   373 & 0 \% \\
		9 & 14 & parking meter   &   16 & 41 \% &    39 & 0 \% \\
		10 & 19 & horse           &   81 & 38 \% &   214 & 0 \% \\
		\hline
		\hline
		71 & 47 & cup             &  128 & 9 \% &  1383 & 1 \% \\
		72 & 27 & backpack        &   54 & 7 \% &   756 & 1 \% \\
		73 & 64 & potted plant    &   62 & 6 \% &   980 & 1 \% \\
		74 & 55 & orange          &   46 & 6 \% &   728 & 1 \% \\
		75 & 80 & toaster         &    1 & 6 \% &    17 & 0 \% \\
		76 & 57 & carrot          &   12 & 2 \% &   631 & 1 \% \\
		77 & 67 & dining table    &   33 & 1 \% &  3105 & 3 \% \\
		78 & 31 & handbag         &    9 & 1 \% &   896 & 1 \% \\
		79 & 37 & sports ball     &    1 & 1 \% &   105 & 0 \% \\
		80 & 89 & hair drier      &    0 & 0 \% &    36 & 0 \% \\
		\hline             
	\end{tabular}
	\caption[Results for the distinct categorical controllability with neighbors]{The top and bottom ten results for the categorical controllability in relation to the 52,107 box captions in the distinct subset. The matches allow here also to include the nearest neighbor words. Nevertheless, box captions were ignored, when their category or one of its nearest neighbor words was already mentioned by the normal caption. The matches are related to the total amount of samples for the category in the distinct subset.}
	\label{table:results_categorical_distinct_k5}
\end{table}


\printbibliography[heading=bibintoc,category=cited]
\printbibliography[title={Further Reading},notcategory=cited]


\end{document}